\documentclass[journal]{IEEEtran}

\usepackage{booktabs,ragged2e}
\usepackage[flushleft]{threeparttable}

\usepackage{graphicx}
\usepackage{algorithm} 
\usepackage{algpseudocode} 
\usepackage{amsmath}
\usepackage{courier}
\usepackage[dvipsnames]{xcolor}
\usepackage{cite}
\usepackage{stfloats}
\usepackage{url}

\usepackage[final]{changes}

\begin{document}

\title{General Hazard Detection}

\author{Stephanie Ng, CP Lim, SueJen Looi, Hendrik Zurlinden, David Nguyen, Lei Wei, Saeid Nahavandi and Hailing Zhou
\thanks{S.~Ng, C.~Lim, S.~Looi, S.~Nahavandi and H.~Zhou were with Swinburne University of Technology, VIC, Australia (email: stephanie13ng@gmail.com, cplim@swin.edu.au, looisuejen@gmail.com, snahavandi@swin.edu.au, hailing.zhou@hotmail.com)}
\thanks{H. Zurlinden was with National Transport Research Organisation (NTRO), Australia (email: hendrik.zurlinden@ntro.org.au)}
\thanks{D. Nguyen was with Google Cloud, California, USA (email: davenguyen@google.com) }
\thanks{L.~ Wei was with Deakin University, VIC, Australia (email: lei.wei@deakin.edu.au )}}


\maketitle

\begin{abstract}
Hazard, as an abstract concept, is typically defined through cognitive-level logical reasoning rather than concrete examples. In contrast, existing hazard detection systems rely on predefined hazard categories and require intensive collection of labelled examples within detection or classification architectures. This approach faces three fundamental challenges when addressing abstract safety concepts: (1) noisy and sparse training data, (2) dynamically evolving definitions that change across contexts and time, and (3) limited generalisation to unseen or novel scenarios. To address these limitations, we present the CompliVision dataset, the first general-purpose hazard dataset designed for rule-based \replaced{compliance assessment, along with a baseline framework for hazard evaluation.}{reasoning, and introduce a baseline reasoning framework.} Our key innovation is decoupling the hazard concept from image-based examples by expressing safety requirements through language-based rules. We ground our approach in authoritative domain regulations and ISO standards to define diverse hazard concepts across multiple domains. The CompliVision dataset comprises 3,006 images spanning traffic, construction, and warehouse environments, with each image annotated for compliance against specific safety rules, accompanied by natural language explanations \replaced{highlighting}{of} the supporting visual evidence. To achieve robust generalisation, we develop an active learning framework \replaced{to more effectively guide and refine vision-language models in assessing hazard compliance}{integrated with vision-language modelling techniques}. While state-of-the-art VLMs demonstrate strong capabilities, they struggle with the fine-grained, context-dependent \replaced{interpretation}{reasoning} required for accurate safety assessment. We proposed a general hazard detection framework to address this limitation which combines LLaVA-based visual reasoning with \replaced{with human-in-the-loop feedback}{active learning}. Experimental results demonstrate the effectiveness of our approach, reducing manual labelling effort by over 65\% while achieving performance comparable to full supervised fine-tuning. Our code and dataset are available at: \url{https://github.com/Haruharu-hub/hazard-project}.
\end{abstract}


\IEEEpeerreviewmaketitle

\section{Introduction}

\IEEEPARstart{S}{afety} management depends on early identification and mitigation of hazards, yet this process often relies on manual inspections by safety managers and is hindered by the dynamic, constantly changing nature of work environments \cite{park2017framework, yang2019inferring, jeelani2021real}. Despite advances in automated monitoring, accidents continue to occur when workers fail to comply with established safety rules, leading to injuries, financial losses, and reputational damage. According to the Safe Work Australia 2025 statistics report \cite{safeworkau}, workplace safety remains a critical national challenge, with 80\% of traumatic injury fatalities and the majority of serious injury claims concentrated in six high-risk industries: agriculture, public safety, transport, manufacturing, health care, and construction. Vehicle incidents alone account for 42\% of all fatal injuries, while falls from height represent another 13\%, underscoring the urgent need for more effective, scalable, and context-aware safety monitoring systems that can proactively identify potential hazards and prevent incidents. 

Existing research in hazard detection often defines hazards using a limited set of predefined rules, such as Personal Protective Equipment (PPE) violations or unsafe behaviours \cite{vukicevic2024systematic, chen2025vision}. While these approaches go beyond generic safe or unsafe labels, the rules are frequently oversimplified and narrowly scoped, failing to capture the full complexity and contextual relevance of hazards in real-world environments. Traditional computer vision approaches, such as object-detection models, can reliably identify safety equipment or specific objects in an image, however, they often struggle to determine whether the observed scene complies with relevant safety standards, particularly when lacking scene-level understanding or when objects appear in diverse contexts, leaving a semantic gap between visual detection and textual safety rules \cite{delhi2020detection, li2022computer, chharia2025safe}.

\begin{figure*}[!t]
\centering
\includegraphics[width=0.95\textwidth]{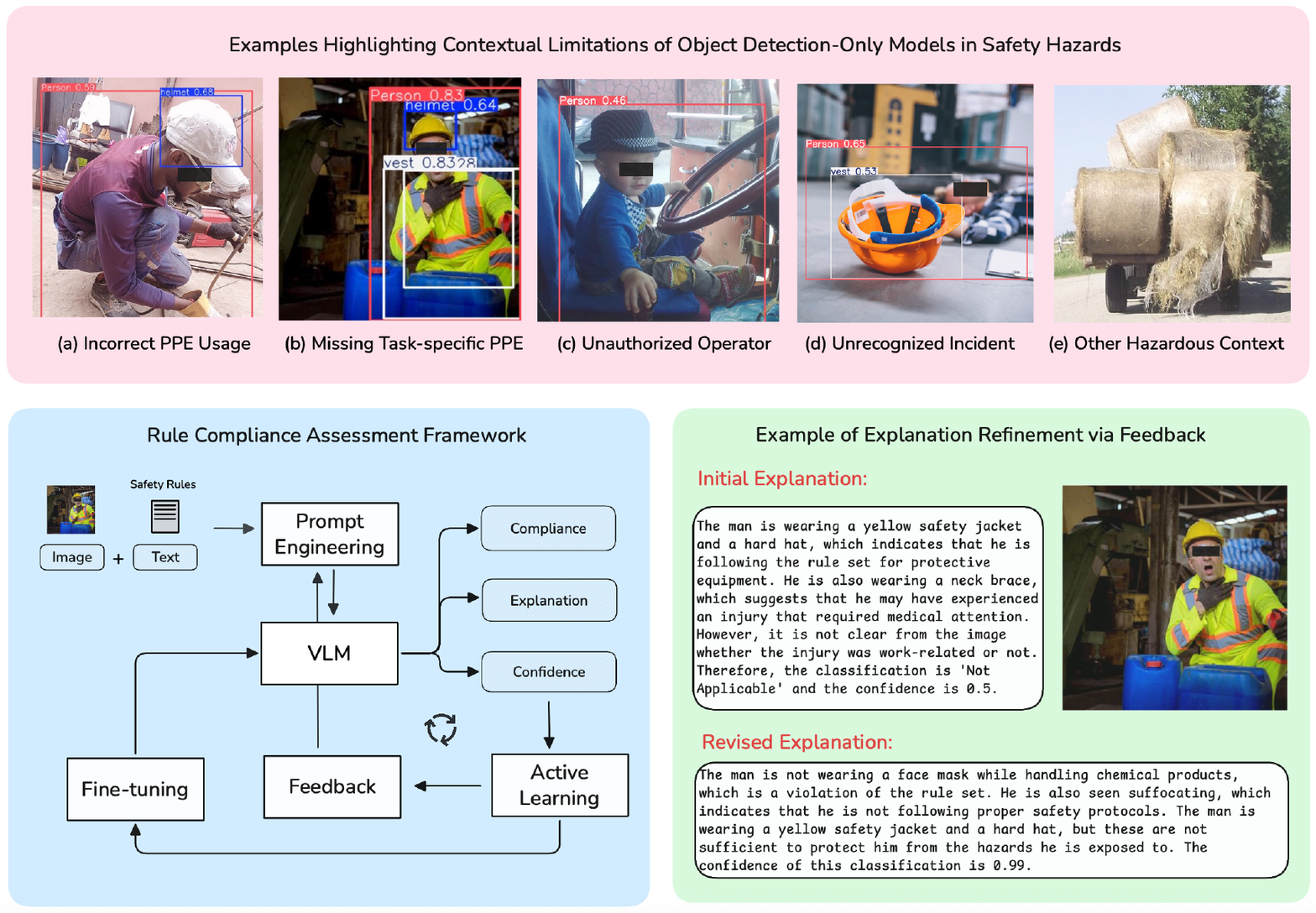}
\caption{Overview of the proposed approach for general hazard detection. Top panel: Key limitations of perception-level object detection models in hazard scenarios. Bottom-left panel: Proposed vision–language model (VLM) pipeline integrating prompt engineering, active learning, human-in-the-loop feedback, and fine-tuning for contextual safety-compliance. Bottom-right panel: Example of a non-compliant PPE scenario demonstrating improved explanation through human-in-the-loop feedback (Image sources: Wikimedia Commons [a, c, e], CC BY-SA 4.0; Vecteezy [b] and Freepik [d], free for use with attribution).}
\label{fig0}
\end{figure*}

\added{Hazards are inherently abstract, heterogeneous, and context-dependent, making them difficult to model using traditional computer vision frameworks as discussed in \cite{our}. Unlike concrete object categories, they cannot be exhaustively enumerated and often evolve over time with changing environmental or societal conditions. The same action may shift from safe to hazardous depending on situational context. For example, crossing during a red signal or standing beneath operating machinery. Safety-critical data further complicates modelling due to its sparsity, noise, and variability across operational settings. These characteristics, including abstraction, temporal variability, contextual sensitivity, and data scarcity, challenge the assumptions of traditional object-detection and classification models (as shown in Fig.~\ref{fig0}), leading standalone models for individual hazards inadequate for this fundamentally abstract task.}

Recent Vision Language Models (VLMs) demonstrate strong multimodal reasoning capabilities, and many approaches use image captioning or visual question answering (VQA) to describe hazards \cite{gil2024zero, ding2022safety}. However, these methods often focus on generating descriptive labels or answering predefined questions rather than explicitly linking visual evidence to the full set of safety rules and standards. Existing benchmarks also lack explicit alignment between visual evidence, natural language safety rules, and compliance judgements, limiting the ability to evaluate whether a model can determine if a safety rule is complied with in a given visual context. In this work, we proposed to solve this challenging task via a fully rule-grounded mechanism which motivates the need for frameworks that explicitly connect visual observations to a broader range of textual rules while assessing compliance in a context-sensitive manner. 

To address this limitation, we propose the CompliVision dataset. In this dataset, each image is explicitly paired with textual compliance rules, reframing hazard detection as a rule compliance testing task. Under this formulation, a hazard naturally corresponds to any instance of rule violation. The model's task is to determine, based on visual evidence, whether each rule is Complied, Violated, or Not Applicable, enabling systematic and interpretable evaluation of potentially unsafe scenarios, even when harm has not immediately materialised. CompliVision encompasses a multi-domain collection of images from construction, warehouse, and traffic environments, each paired with domain-specific rules. Image–rule pairs are annotated through a human-in-the-loop feedback process, where a VLM generates initial predictions that are subsequently verified and corrected by human experts. This approach combines high-quality annotations with scalable automated reasoning, supporting reliable evaluation and model training across multiple safety domains. The key contributions are listed as follows.
\begin{itemize}
    \item \textbf{CompliVision Dataset}: The first multi-domain hazard dataset provides visual scenarios and authoritative regulations, specifically with 3,006 images and 54 safety standards across traffic, construction, and warehouse domains.
    
    \item \textbf{\replaced{Rule Compliance Assessment Framework}{Rule-Based Reasoning Framework}}: A systematic approach that formulates hazard detection as a rule-compliance assessment task, decoupling hazard concepts from predefined object categories and expressing them through language-based regulatory standards to address the challenges of abstraction, contextual variation, temporal evolution, and data scarcity.
    
    \item \textbf{Scalable Active Learning Mechanism}: An efficient training framework integrating VLM-based reasoning with expert validation through disagreement-based sample selection and confidence-calibrated pseudo-labelling, achieving over 65\% reduction in annotation effort.
\end{itemize}

\section{Related Work}

\subsection{Computer Vision for Hazard Detection}

Hazard detection systems have been applied across a wide range of domains. Recent research has leveraged computer vision and deep learning techniques, particularly YOLO-based architectures, for real-time hazard detection across construction and infrastructure environments. Traditional methods primarily focus on detecting PPE, environmental hazards, and infrastructure degradation. 

YOLOv4 has been used to monitor worker helmet compliance by incorporating human body models to ensure correct helmet-worker correspondence \cite{yunyun2021detection}. Similarly, automated pipelines combining worker localisation, semantic segmentation, and hazard detection have been applied for both static and dynamic hazards, supporting real-time monitoring \cite{jeelani2021real}. Expanding this scope, methods have also addressed helmet and mask detection with identity verification, employing YOLOv5 for PPE detection and Facenet + SVM for masked-face recognition \cite{kwak2023detection}. YOLO-based CNNs have detected road surface degradation (potholes, manholes, blurred markings) and applied semantic segmentation for pixel-level localisation \cite{pena2020real}. Enhanced YOLOv4 networks detect small sparks under variable conditions and provide real-time alerts, demonstrating applicability for smart city monitoring \cite{avazov2021fire}. These approaches highlight the efficacy of YOLO and CNN-based object detection for real-time hazard recognition. However, traditional models often lack adaptability to evolving safety regulations, motivating integration with knowledge-based and reasoning systems.

Another line of research focuses on neural-symbolic and modular frameworks for interpretable visual reasoning. Neural Module Networks (NMNs) and variants such as the Meta Module Network (MMN) dynamically construct reasoning modules based on input queries, enabling scalable multi-hop reasoning \cite{chen2021meta}. MMNs can represent previously unseen functions via embeddings, addressing limitations of traditional NMNs. The NS-VQA framework integrates structured scene representations with symbolic program execution, providing robust, interpretable, and data-efficient reasoning \cite{yi2018neural}. Similarly, the GQA dataset introduces a visual reasoning benchmark for compositional question answering, leveraging object, attribute, and relational information from Visual Genome scene graphs \cite{hudson2019gqa}. Program generator and execution engine approaches explicitly model reasoning steps, outperforming black-box models on benchmarks such as CLEVR \cite{johnson2017inferring}. While neural-symbolic frameworks offer interpretable and compositional reasoning, they do not fully integrate visual observations with textual safety rules or evaluate compliance in context. This gap underscores the need for vision-language models capable of multimodal interpretation and generation.

Beyond purely vision-based methods, structured knowledge representations such as knowledge graphs (KGs) and ontology-based models enable automated reasoning over dynamic and heterogeneous safety information. KGs and ontologies encode hazards, relationships, and evolving regulations, supporting accurate detection and compliance even as rules change \cite{xiao2023review}. KGs also integrate textual and visual data, using entity extraction with BERT-BiLSTM-CRF and visual feature extraction via YOLOv5-FastPose, enabling multi-dimensional queries and knowledge visualisation \cite{fang2020knowledge}. Reviews of construction safety KGs highlight systematic development steps, including scope identification, ontological modelling, data extraction, and KG completion, and discuss challenges for real-world adoption \cite{wu2023construction}. In autonomous driving, graph-based methods model road users, static objects, and traffic rules to support relational reasoning for hazardous event detection. These methods can be rule-based, probabilistic, or learning-driven, providing structured insights into complex environments \cite{zhong2020hazard}. Complementary safety assurance frameworks ensure that perception modules based on deep neural networks adhere to standards such as ISO 21448 and ISO PAS 8800 \cite{abrecht2024deep}. \replaced{While knowledge-based approaches improve interpretability and regulatory alignment, they typically rely on query-driven reasoning mechanisms and depend on carefully curated knowledge graphs, which are often costly to construct and maintain. Moreover, ontology-based representations may introduce structural constraints that limit representational flexibility}{While knowledge-based approaches enhance interpretability, adaptability, and regulatory alignment, they rely on query-driven mechanisms and require comprehensive KG databases; moreover, ontology-based representations impose structural constraints that limit representational flexibility.}

\subsection{Hazard Analysis Datasets}

Prior research has produced extensive datasets and benchmarks for hazard detection across construction, traffic, and industrial domains. However, most existing work remains centred on object-level detection, offering limited support for rule-centric reasoning, contextual interpretation, or multimodal understanding. In the construction domain, numerous datasets target worker safety, particularly around personal protective equipment (PPE) such as helmets, vests, and goggles \cite{dalvi2025construction}. Broader scene-level datasets such as SODA capture categories including persons, materials, machines, and layout elements \cite{duan2022soda}, while MOCS focuses on the detection of moving objects, including workers and vehicles operating on construction sites \cite{xuehui2021dataset}. To mitigate the limitations of retraining models for each new object category, ConstructionSite10k introduces semantic features designed to enhance VLM capabilities, supporting image captioning, multiple-choice reasoning, question answering, and visual grounding based on a curated selection of MOCS images \cite{chen2025large}.

In the traffic domain, CLEVRER proposes a synthetic benchmark for causal and counterfactual reasoning over collision events \cite{yi2019clevrer}, while DeepAccident advances early accident forecasting through multimodal V2X sensor data with multi-agent trajectories and dense interaction patterns \cite{wang2024deepaccident}. Although these datasets capture high-risk scenarios, they do not encode domain rules such as right-of-way, pedestrian-yield requirements, or safe-following guidelines. Complementing these efforts, the WTS dataset emphasises fine-grained spatio-temporal understanding of pedestrian behaviour in traffic scenes \cite{kong2024wts}.

In the industrial and manufacturing domain, datasets such as SH17 address PPE detection in factory environments \cite{ahmad2024sh17}, while other work focuses on early detection of specific hazards, particularly fires and smoke, across indoor and outdoor settings \cite{elhanashi2025early}. Together, these datasets provide strong foundations for perception and event-level analysis but fall short of supporting generalised, rule-aware hazard reasoning across domains. In contrast, our dataset aims to enable domain-independent, rule-aware hazard reasoning by leveraging in-context learning to interpret arbitrary safety rules and apply them consistently across traffic, construction, and industrial scenarios.

\subsection{Vision Language Models}

Recent advances in vision–language models (VLMs) have enabled large language models to perform increasingly sophisticated visual reasoning tasks by combining image understanding with linguistic inference \cite{liu2023visual}. Chain-of-thought prompting has been shown to improve multi-step reasoning in language models \cite{wei2022chain}, and similar principles have been extended to visual domains. For example, LLaVA-CoT structures VLM outputs into sequential phases of scene summarisation, visual interpretation, logical reasoning, and final conclusions to reduce hallucination and promote transparent, systematic reasoning \cite{xu2025llava}. Research on hazard understanding increasingly incorporates VLMs, drawing on image captioning, VQA, and zero-shot scene interpretation as mechanisms for extracting safety-relevant cues from images \cite{chen2025large, chen2025vision, gil2024zero}. However, these approaches typically optimise for general visual understanding rather than explicit safety-rule interpretation. In practice, safety policies are dynamic, context-dependent, and often organised hierarchically, meaning that correct hazard assessment requires aligning model outputs with domain-specific rules and human expectations about acceptable risk. Existing systems rarely incorporate structured feedback loops to encode preferences, resolve ambiguities, or ensure rule-consistent judgements, whereas human-in-the-loop strategies can guide or correct model reasoning \cite{klein2024interactive, kandiyana2024active}, which motivates this study.

\section{The CompliVision Dataset}
\subsection{Image Collection}
To support rule-grounded hazard assessment, we introduce the CompliVision dataset. The dataset is designed to capture diverse safety-critical scenarios across three domains: construction, warehouse, and traffic.

Our data collection process began with a targeted search of open-source stock image repositories (e.g., Vecteezy, Unsplash, Pexels) using five domain-specific keywords per category, each corresponding to a distinct safety rule type. This keyword-based retrieval strategy naturally provides an initial binary compliance label, as images typically depict either rule compliance (0) or a clear violation (1).

\begin{table*}[!b]
\begin{threeparttable}
\caption{Summary of CompliVision dataset}
\label{tab1}
\setlength{\tabcolsep}{0.7\tabcolsep}
\begin{tabular*}{\textwidth}{@{\extracolsep{\fill}} ll ccccccc}
\toprule
     Domain & Safety Regulation (Subrules) & 
     \multicolumn{3}{c}{Image Statistics}  & \multicolumn{3}{c}{Annotation Statistics} & Image-Rule Samples \\ 
\cmidrule{3-5} \cmidrule{6-8} & 
    & Processed & Label 0/1 & Total Images & Complied & Not Applicable & Violated  \\
\midrule
     Traffic & Driving Distraction (4)  & 207 & 100/107 & & 112 & 788 & 107 & \\
    & Pedestrian Crossing (3) & 200 & 100/100 & & 136 & 760 & 111 & \\
    & Road Condition (3) & 200 & 100/100 & 1007 & 642 & 211 & 154 & 5035 \\
    & Traffic Rules (10)  & 200 & 100/100 & & 385 & 453 & 169 &\\
    & Vehicle Load (2)  & 200 & 100/100 & & 169 & 735 & 103 &\\
\midrule
\addlinespace
     Construction & Crane Use(3) & 200 & 100/100 & & 127	& 763 & 109 & \\
    & Fire Risk (2) & 200 & 100/100 & & 121 & 771	& 107 & \\
    & Ladder Use (5) & 199 & 100/99 & 999 & 105 & 795 & 99 & 4995\\
    & Protective Equipment (3) & 200 & 100/100 &  & 430 & 282 & 287 & \\
    & Scaffolding Risk (4) & 200 & 100/100 & & 128 & 733 & 138 &\\
\midrule   
\addlinespace
      Warehouse & Ergonomic Lifting(2) & 200 & 100/100 & & 137	& 717 & 146 &\\
    & Forklift Use(3) & 200 & 100/100 & & 124 & 775 & 101 & \\
    & Ladder Use(4) & 200 & 100/100 & 1000 & 102 & 788 & 110 & 5000 \\      
    & Protective Equipment(3) & 200 & 100/100 & & 319 & 211 & 470 & \\
    & Surface Condition(3) & 200 & 100/100 & & 530 & 288 & 182 &\\
\addlinespace
\midrule
     Total & & & & 3006 & 3567 & 9070 & 2393 & 15030\\
\bottomrule
\end{tabular*}
\smallskip
\scriptsize
\end{threeparttable}
\end{table*}

\subsection{Pre-processing}
All images in the CompliVision dataset underwent a standardised preprocessing pipeline to ensure consistency across domains and suitability for VLM input requirements. First, duplicate and near-duplicate images were identified and removed using perceptual hash matching, ensuring dataset diversity and reducing redundancy. Each remaining image was then resized to 336 × 336 pixels to match the input resolution of the VLMs used in this study. To maintain spatial consistency and focus on the most informative region, a center crop operation was applied after resizing.

To ensure legal and contextual relevance, each rule set in the CompliVision dataset is derived from authoritative safety legislations and standards, primarily including OSHA regulations for workplace safety and the Road Safety Road Rules 2017 for traffic environments (See Appendix A). During the rule curation process, we aim to preserve the original legislative wording as much as possible to maintain semantic precision and regulatory authenticity. However, to reduce linguistic ambiguity and ensure that each rule can be unambiguously interpreted by both human annotators and language models, we remove clauses containing conditional modifiers such as “except” or “unless”.

This filtering process ensures that each rule expresses a clear, direct compliance condition that can be consistently evaluated against visual evidence. Subsequently, each image in the dataset is explicitly paired with five domain-specific rule sets, corresponding to distinct safety categories within construction, warehouse, and traffic domains. These image-rule pairs serve as the foundation for multi-rule compliance classification based on visual cues.

In total, the CompliVision dataset contains 3,006 images across three safety-critical environments: traffic (1,007 images), construction (999 images; one excluded due to duplication), and warehouse (1,000 images). Table \ref{tab1} summarises the dataset composition and domain distribution.

\subsection{Manual Compliance Labeling}
Before introducing model-assisted label refinement, an initial round of manual annotation was conducted to establish high-quality baseline labels used for all experiments. Human annotators evaluated each image–rule pair and assigned one of three compliance labels based on the observable evidence in the image. To maintain consistency across annotations, all decisions were guided by a predefined annotation schema (See Appendix B) that defines observable evidence requirements and handling of ambiguity. 

\begin{figure*}[!h]
\centering
\includegraphics[width=0.95\textwidth]{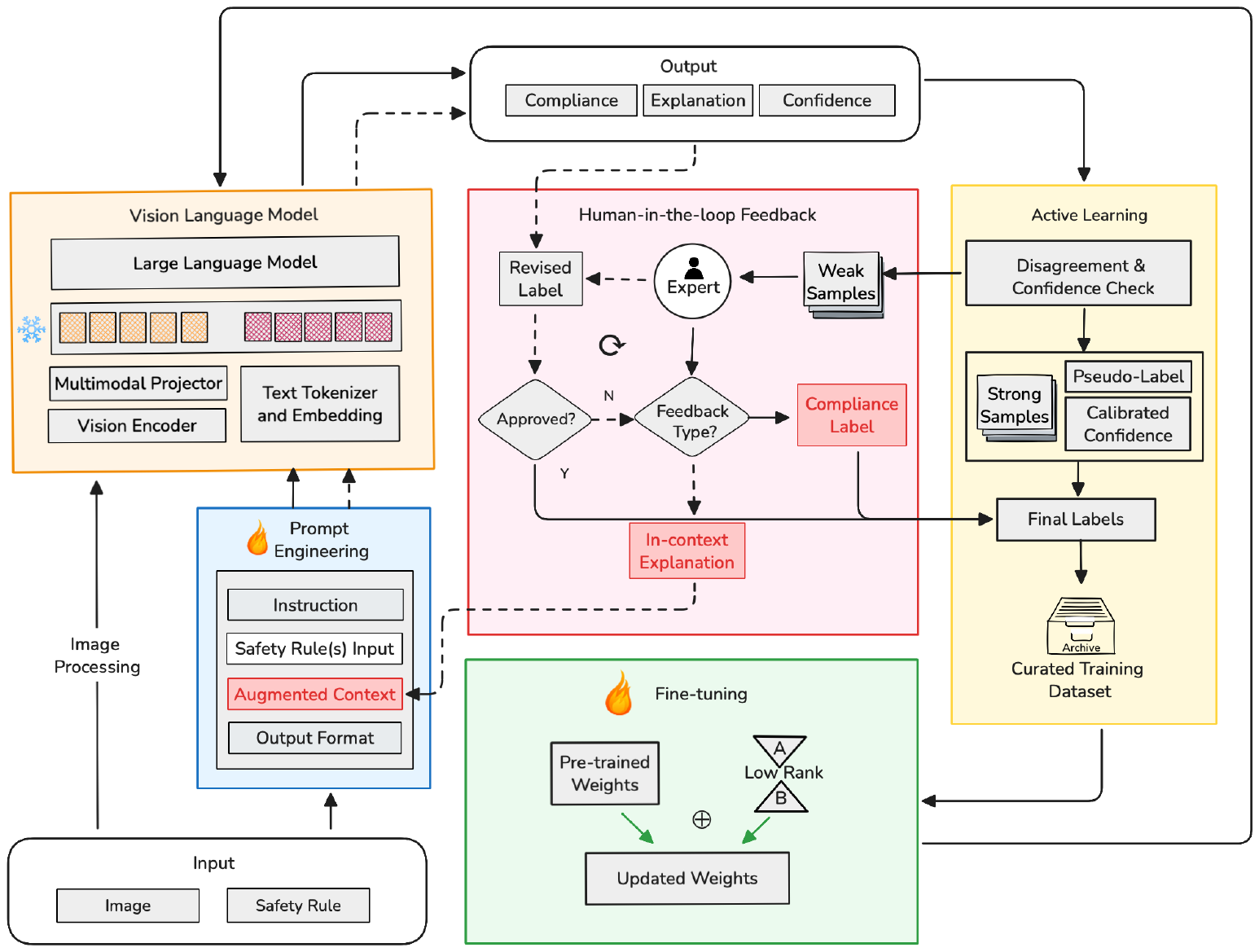}
\caption{Illustration of the proposed framework in details.}
\label{fig1}
\end{figure*}

\section{Methodology}

We formulate the general hazard detection problem as a rule-based compliance reasoning task. Given an input image $I$ depicting a scenario and a set of safety rules $R = \{r_1, r_2, \ldots, r_n\}$ where $n$ denotes the total number of rules, the objective is to assess compliance of the observed scene with respect to each rule in $R$. For each image-rule pair $(I, r_i)$, the task consists of two steps: (1) determining whether rule $r_i$ is applicable to the scenario depicted in $I$, and (2) if applicable, assessing whether the visual evidence indicates compliance or violation of $r_i$. The output for each pair is a compliance label $c_i \in \{\text{Complied}, \text{Violated}, \text{Not Applicable}\}$.

Figure \ref{fig1} shows the technical details of our proposed framework. The process begins with an input pair consisting of an image and a corresponding safety rule. The image is preprocessed and cached for the vision encoder, while the rule set is incorporated into a curated prompt template before being passed to the VLM for tokenisation and embedding. The parameter-frozen VLM then generates a response that is post-processed into three structured outputs: Compliance, Confidence, and Explanation. These outputs enable targeted feedback interactions: low-confidence or ambiguous cases are flagged for human verification or correction, supporting active learning and iterative refinement of model behaviour. \replaced{The next section details five key components of the system, including prompt engineering, VLM, active learning, human-in-the-loop feedback, and fine-tuning.}{The next section details key components of the system, including \textcolor{blue}{VLM}, prompt engineering, active learning, and feedback.}

\subsection{Prompt Engineering} 
\deleted{We need a paragraph to introduce prompt engineering here to show its background/motivation, ensuring the completeness of the article.. something like this. The safety rules can not directly used as the inputs of VLM engine. We need to transform them into prompts. As analyzed in , prompts are essential for translating general users inputs (i.e. safety rules in our work) into structured instructions of large language models (LLMs). The design of prompts - prompt engineering - is crucial for large language models. Well-crafted prompts enable task adaptation, proper model reasoning, and improve output consistency without requiring additional training.}

\added{Prompt engineering plays a crucial role in guiding VLMs to accurately interpret and act on inputs. Directly feeding safety rules into a VLM often leads to ambiguous or inconsistent behaviour, particularly for novel or domain-specific tasks. To address this, we leverage the language understanding capabilities of LLM to enable zero-shot task adaptation. As discussed in} \cite{sahoo2024prompt} \added{, prompts, or carefully designed instructions, are essential for transforming general user inputs, including safety rules, into structured guidance that VLMs can reliably follow. Well-crafted prompts not only facilitate task adaptation and guide model reasoning but also improve output consistency without requiring additional model training.}

Building on this, we designed three types of prompts, as follows.

\begin{itemize}
    \item \textbf{Task-focused variants (T1-T4)}: Minimal prompts that state the compliance prediction task without additional guidance. T1 instructs the model to classify the image with respect to a given rule. T2 constrains the model to respond using exactly one predefined compliance category. T3 is derived from T2 by changing a single keyword, replacing \emph{according to} with \emph{against}. T4 extends T3 by instructing the model to analyse the image before responding with one of the compliance labels.
    
    \item \textbf{Classification-focused variants (C1-C4)}: A set of progressively structured prompts that emphasise output consistency and decision clarity. C1 provides a base classification prompt that explicitly constrains model responses to a standardised output format, using single-token labels in JSON-style to enhance parsing reliability. Subsequent variants (C2-C4) incorporate increasingly directive instructions that emphasise rule applicability, confidence estimation, and reflective evaluation ("think-before-answering") prior to producing a final classification. 

    \item \textbf{Explanation-focused variants (E1-E2)}: Prompts designed to generate a brief justification alongside compliance label. E1 represents an extension of C4 that instructs the model to provide a concise rationale supporting its classification, enabling interpretability analysis. E2 represents a simpler variant that extends C1, similarly requesting short rationales to facilitate interpretability evaluation under minimal prompting conditions.

\end{itemize}

This setup allows us to systematically measure how prompt type and design choices influence the VLMs' ability to reason across domains and rules, and to identify prompt variants that achieve the best balance between prediction accuracy and interpretability. Complete prompt templates are provided in Appendix C. The optimal prompt design is selected based on empirical performance using LLaVA as our base model, with detailed analysis presented in Section VI-A. 

\subsection{Visual Language Model}

\deleted{To be consistent and keep the description complete, better to cover a bit deteails on VLM here}

We employ a VLM as the multimodal backbone that executes the prompt described in the previous section. A VLM is a multimodal neural architecture that enables joint processing of visual and linguistic inputs by integrating a vision backbone with a large language model through learned cross-modal alignment \cite{liu2024large}. 

Given an input image $I$ and a text sequence $T$, where $T$ consists of the curated prompt template along with the relevant safety rule set $R$, a VLM consists of the following components to process the visual and textual inputs and generate task-specific outputs:

\begin{itemize}
    \item \emph{Visual Encoder} $f_v(\cdot)$: a vision tower that extracts high-level feature embeddings from the input image $I$. Typically implemented using a convolutional neural network or a Vision Transformer (ViT), it converts the input image into a sequence of visual embeddings that capture objects, spatial structure, and semantic content. Then, a vision resampler may further refine or downsample the visual embeddings to control the number of visual tokens.

    \item \emph{Multimodal Projector} $g(\cdot)$: maps visual features into a representation space compatible with the language model’s token embeddings, facilitating cross-modal alignment. 

    \item \emph{Text Tokenizer and Embedding} $f_t(\cdot)$: converts the input text $T$ into a sequence of discrete tokens and corresponding continuous embeddings.

    \item \emph{Large language model} $f_{\mathrm{LLM}}(\cdot)$: a transformer-based decoder with masked Multi-Head Attention layers and a language modeling (LM) head. Masked attention ensures tokens attend only to previous tokens (including visual embeddings), preserving causality, while the LM head maps hidden states to vocabulary logits. This enables the VLM to generate coherent multimodal outputs grounded in visual context.
\end{itemize}

During the initial prompt engineering stage, all parameters are frozen preserve pre-trained visual-language alignment. In this study, we evaluate four state-of-the-art VLMs, as detailed in Section V-A Relevant experimental results are presented in Section VI-B. 

\deleted{Different state-of-the-art VLMs are evaluated in our experiments where the detailed results are shown in Section V-A and VI- B. In this work, an additional component such as multimodal fusion layers, instruction-tuning modules, and task-specific prediction heads is incorporated to enhance cross-modal alignment, reasoning capability, and robustness for our tasks.}

\subsection{Active Learning}

\deleted{Similarly, we need a paragraph of rational analysis of using AL via 1) introducing the basis of active learning for completeness, 2) describing the importance/motivations of using active learning.}

General hazard detection task is often challenged by noisy and sparse data. To substantially reduce data requirements and annotation effort, an active learning-based training pipeline is explored. Active learning iteratively trains a model by selecting the minimal number of samples needed to achieve maximal performance \cite{ren2021survey}.

The critical questions in active learning are: (1) how to select samples for manual annotations (i.e., how to identify informative samples) and (2) what types of annotations are required for the selected samples to maximise learning efficiency. 

To address the first question, a dual-criteria sampling algorithm is developed to determine which samples to choose at each iteration. Algorithm \ref{algo_1} outlines the proposed active learning procedure, covering initialisation and iterative active learning, including sample selection, human annotation, and pseudo-labelling steps. The active learning process is initialised using two prompt variants (C1 and C4) as complementary probes for sample selection. The C1 prompt serves as the baseline classifier, while the C4 prompt incorporates additional reasoning cues, applicability checks, and an explicit confidence output. These prompt outputs are used as our sample selection criteria. Informative samples are identified based on model uncertainty and disagreement (i.e., low confidence score and variation in predictions across prompts), which constitute the weak samples. The remaining unlabelled samples are proceeded using the pseudo-labelling technique. This setup allows us to directly measure the trade-off between annotation cost and model performance. Specifically, each sample is categorised as follows: 

\begin{itemize}
    \item \textbf{Weak samples for human annotation}: Instances where model predictions across rules are inconsistent or exhibit low confidence (below the distributional maximum). These are considered informative samples, as they are likely to reduce model uncertainty. Human labels are requested only for these weak samples, representing the costly portion of the annotation process.

    \item \textbf{Strong samples via pseudo-labelling}: Instances where predictions are consistent and the model exhibits high confidence. These samples are automatically pseudo-labelled using the model's predicted rule-level compliance labels. To calibrate reliability, pseudo-label confidence scores are adjusted according to their embedding distance from the nearest human-labelled reference centroid. Confidence is preserved or boosted when the nearest centroid's rule-label pair matches the pseudo-label prediction, and downscaled otherwise, reflecting reduced trust in labels inferred without human verification.
    
\end{itemize}

In contrast to the traditional fully fine-tuning-based training pipeline, which assumes that human annotations are available for the entire dataset, our active learning pipeline progressively queries human labels only for the most informative samples, thereby significantly reducing both the amount of training data required and the annotation effort.

\begin{algorithm}
\caption{: AL with Disagreement Evaluation and Confidence Calibration for Pseudo-Labelling} \label{algo_1}
\begin{algorithmic}[1]


\Statex \textbf{Inputs}: Train set $D_{\text{Train}}$, Validation set $D_{\text{Val}}$,  Test set $D_{\text{Test}}$
\Statex \textbf{Outputs}: Fine-tuned model $M$
\Statex \hrulefill

\Statex \textbf{Phase 1: Initialisation} 
\State Select a base vision-language model $M_0$ and two prompt templates for disagreement evaluation.
\State Split Train set $D_{\text{Train}}$ into initial training pool $D_{\text{Init}}$ with $N$ samples and remaining training pool $D_{\text{Add}}$.
\For{each sample $x$ in $D_{\text{Init}}$}
    \State Predict rule-compliance label $\hat{y}$ and confidence $C(\hat{y})$ using $M_0$.
    \State Categorise samples as \textit{weak} (disagreement or low-confidence cases) or \textit{strong} (otherwise).
\EndFor
\State Obtain human annotations for weak samples: $D_{\text{Weak}} \gets \{(x, y, C)\}$, where $C = 1$.
\State Assign pseudo-labels to strong samples: $D_{\text{Strong}} \gets \{(x, \hat{y}, C)\}$, where
\[
C = C_{\text{base}} \times \alpha\Big(\hat{y}, \arg\min_i \text{dist}(\text{emb}_x, \text{emb}_{\text{centroid}_i})\Big)
\]
\State Update initial training set: $D_{\text{Init}} \gets D_{\text{Weak}} \cup D_{\text{Strong}}$.
\State Train model $M$ using $D_{\text{Init}}$, validating on Validation set $D_{\text{Val}}$ for early stopping.
\Statex \hrulefill
\Statex \textbf{Phase 2: Iterative Active Learning}
\While{annotation budget $B$ is not exhausted}
    \For{each new sample $x^*$ in $D_{\text{Add}}$}
        \State Predict rule-level compliance labels $\hat{y}$ and confidence $C(\hat{y})$ using $M$.
    \EndFor
    \State Select uncertain or high-disagreement samples for human annotation.
    \State Assign pseudo-labels to remaining samples.
    \State Update accumulated $D_{\text{Init}}$ and retrain $M$.
\EndWhile
\State Evaluate final fine-tuned model $M$ on Test set $D_{\text{Test}}$.
\end{algorithmic}
\end{algorithm}

\begin{figure}[!ht]
\centering
\includegraphics[width=0.45\textwidth]{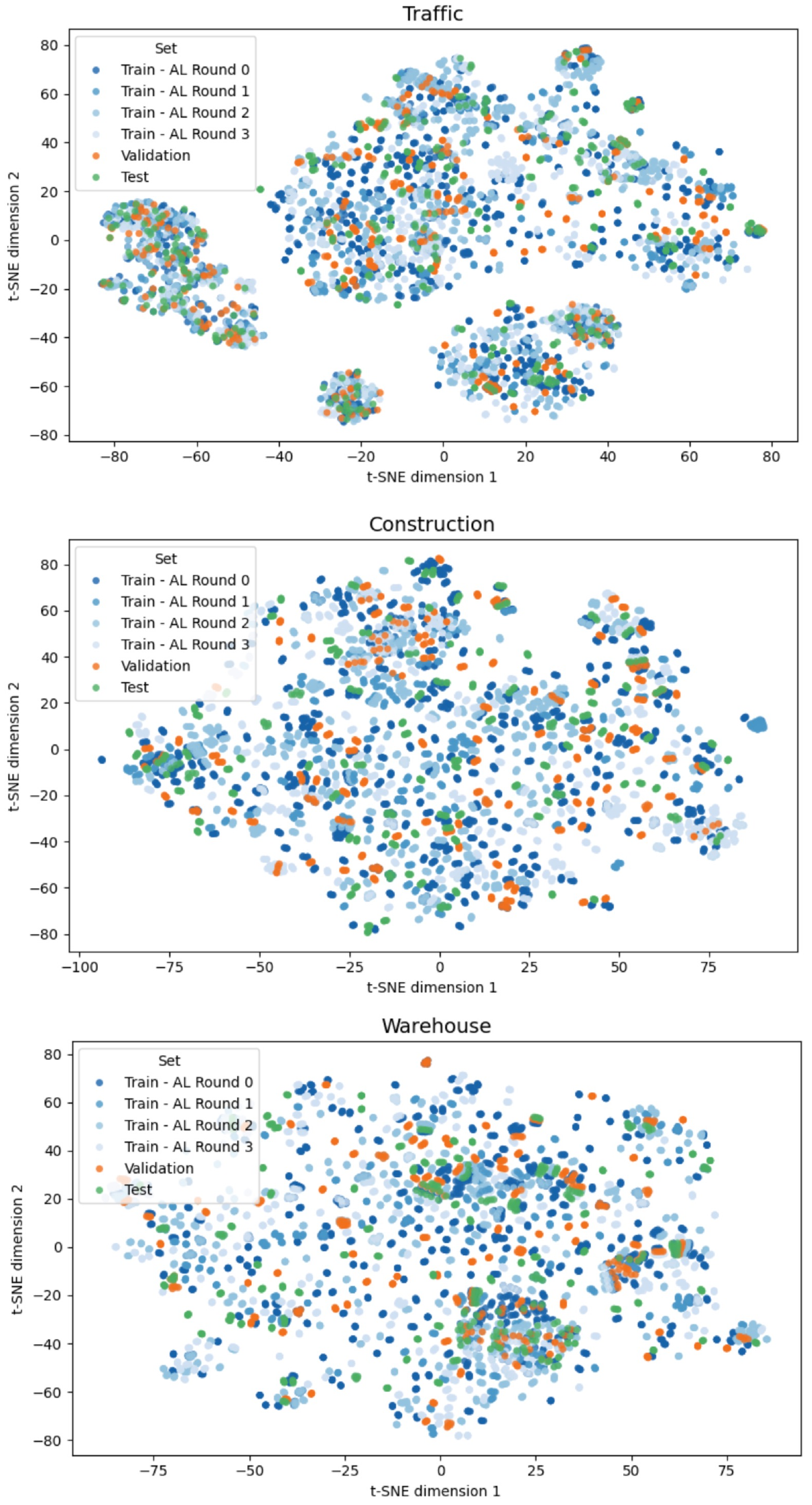}
\caption{t-SNE Visualization of Embeddings for Training, Validation, and Test Data across Active Learning Rounds.}
\label{fig2}
\end{figure}

Figure \ref{fig2} illustrates the t-SNE projection of embeddings for training, validation, and test data across the active learning rounds, showing the distribution of selected samples in the feature space.

\subsection{Human-in-the-loop Feedback}

To address the second question in active learning, we explore two forms of feedback: compliance label and in-context explanation. The feedback is used as augmented context to significantly enhance the prompt engineering component showing in Fig.~\ref{fig1}.

\subsubsection{Compliance Label}
In this mode, annotators provide authoritative labels (i.e., complied, not applicable, or violated) for weak samples. The revised labels are directly incorporated into the training set for fine-tuning, enabling supervised updates and ensuring that the model progressively aligns with domain-specific safety expectations.

\subsubsection{In-context Explanation}
\begin{figure}[!ht]
\centering
\includegraphics[width=0.45\textwidth]{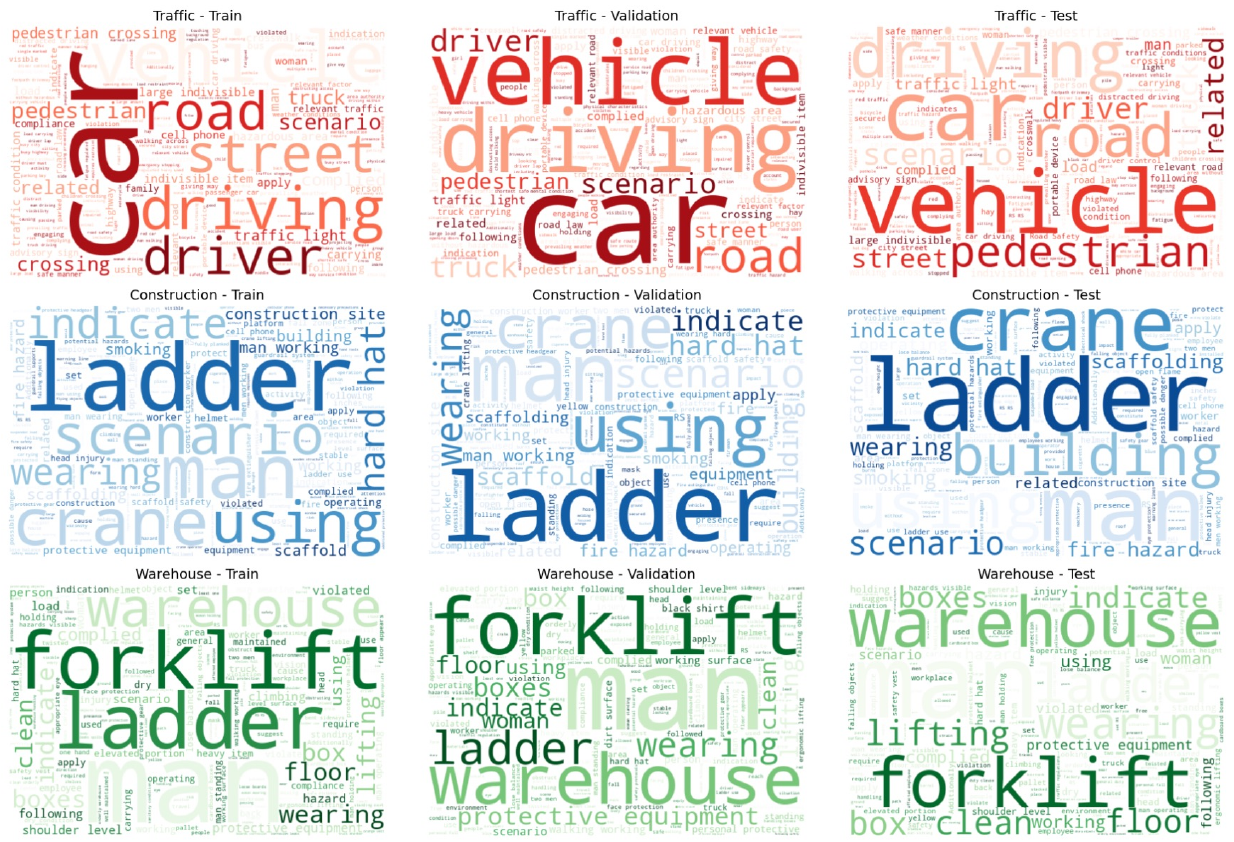}
\caption{Word Clouds of Generated Explanations for Training, Validation, and Test Data.}
\label{fig3}
\end{figure}

In-context explanation allows annotators to provide natural-language feedback that clarifies rule conditions, highlight relevant visual cues, and explain the rationale behind the correct compliance judgement. Figure \ref{fig3} presents word clouds of the in-context explanation for the training, validation, and test sets, highlighting the most frequently mentioned objects and visual elements. The detailed implementation is outlined in Algorithm \ref{algo_2}, which has two main phases: an initialisation stage and an iterative feedback loop.

\added{The initialisation phase is to generate an initial explanation for every sample via the $E1$ prompt template (see Appendix C-C)). It is then iteratively augmented during the feedback loop phase. The feedback loop phase phase is to identify weak samples and then incorporate proper human feedback for the prompt refinement as shown in the red box of Fig.~\ref{fig1}. To have a proper feedback, for each sample, the most relevant prior feedback is firstly retrieved based on embedding similarity and added to the current context, if available. This step allows the model to reference explanations humans provided for similar past cases and guides the model to generate improved predictions and explanations. If the sample is still classified as weak and remains uncertain, additional human feedback is then collected to guide corrections. The prompt is temporarily updated with this in-context explanation and added permanently to the prompt only if it remains consistent with previously seen samples; otherwise, the update is discarded. All seen samples are tracked to ensure stability, preventing the model from “forgetting” previous guidance while iteratively refining the prompt. This process ensures that only reliable human feedback is incorporated for future inferences.}

\begin{algorithm}
\caption{: In-Context Feedback} \label{algo_2}
\begin{algorithmic}[1]

\Statex \textbf{Inputs}: $D_{Train}$ with samples $(x_i, y_i)$, where $x_i$ represents an image and $y_i$ represents the corresponding classification labels. 
\Statex \textbf{Outputs}: In-context feedback $\psi$
\Statex \hrulefill

\Statex \textbf{Phase 1: Initialisation} 
\For{each sample $(x, y)$ in $D_{\text{Train}}$}
    \State Generate initial explanation $E_0$.
\EndFor
\State Initialise context $\psi = \text{BasePrompt}$; Seen samples $S \gets \emptyset$ ; Feedback memory $F \gets \emptyset$.
\Statex \hrulefill
\Statex \textbf{Phase 2: Feedback Loop} 
\For{each batch $b$ in $D_{\text{Train}}$}
    \State Compute and store image embeddings $e_x$ 
    \State $e_x = \text{Embed}(x)$ for all $x$ in $b$.
    
    \For{each sample $x_i$ in $b$}
        \State Retrieve most relevant feedback $f^*$
        \State $f^* = \text{Retrieve}(F, e_x, \tau)$ $\tau$ is a threshold
        \State Update context $\psi$
        \State Run model $M$ with current context $\psi$,
        \State Get the response: $(\hat{y}_i, \hat{E}_i) \gets M(x_i, \psi)$.
        \If {selected as weak sample}
            \State Collect human feedback $f_i$
            \State Generate revised response $(\hat{y}_i^*, \hat{E}_i^*)$.
            \If{feedback not accepted or is skipped} 
                \State \textbf{continue} to next sample.
            \EndIf
            \State Propose updated context $\psi'$ 
            \State $\psi'\gets \text{AugmentedContext}(\psi, f_i)$.
            \State Check stability on seen samples.
            \If{stable} 
                \State Update context: $\psi \gets \psi'$.
                 \State Add $(x_i, f_i)$ to feedback memory $F$
            \Else
                \State Discard $\psi'$ (revert update).
            \EndIf
        \EndIf
        \State Seen samples $S \gets S \cup \{(x_i, y_i)\}$.
    \EndFor
\EndFor
\end{algorithmic}
\end{algorithm}

\subsection{Fine-tuning}

To efficiently adapt the base VLM for rule-grounded compliance assessment, we employ Low-Rank Adaptation (LoRA) fine-tuning \cite{hu2022lora}. LoRA enables parameter-efficient training by injecting low-rank trainable matrices into selected model's layers while keeping the original pre-trained weights frozen. This strategy substantially reduces the number of trainable parameters and GPU memory usage, enabling fine-tuning on limited data without over-fitting or catastrophic forgetting. \added{In our setup, LoRA adapters are applied primarily to the transformer layers of the LLM. A very small subset of Linear layers in the vision tower and multimodal projector also receive adapters. The remaining modules, including LM head, are frozen. Specifically, out of the total parameters:}

\begin{itemize}
    \item Vision Encoder: 1.2M / 305M trainable (0.39\%)
    \item Multimodal Projector: 0.1M / 21M trainable (0.51\%)
    \item LLM: 20M / 6.6B trainable (0.30\%)
    \item Other: 0 / 131M trainable (0.0\%)
\end{itemize}

\added{The total trainable parameters across all modules amount to 21.3M out of 7.1B (0.30\%). This demonstrates that the vast majority of the model remains frozen, and learning occurs primarily through the small, parameter-efficient LoRA adapters, preserving the pre-trained visual-language alignment while allowing task-specific adaptation.}

\section{Experiments Setup}

\subsection{Baseline Models}

\added{To evaluate the effectiveness of our approach, we consider three baseline settings: (1) zero-shot prompt selection, (2) zero-shot benchmarking across VLM architectures, and (3) fully fine-tuned VLMs. These baselines provide a spectrum of performance references, from no task-specific training to complete adaptation on the full training set.}

\subsubsection{Zero-shot Benchmarking Across Prompt Variants}

We first evaluate the best-performing prompts identified during validation set experiments. These prompt-engineered templates are applied to the base LLaVA VLM without any task-specific training. By comparing three prompt variants: task-focused, classification-focused, and explanation-focused, we evaluate how well each strategy generalises across multiple rules and domains. This allows us to identify the prompts that are most effective for downstream compliance prediction task.

\subsubsection{Zero-Shot Benchmarking Across VLM Architectures}

Using the selected prompts, we then benchmark several state-of-the-art VLMs, including LLaVA \footnote[1]{llava-hf/llava-1.5-7b-hf} \cite{liu2023visual}, LLaVA-Next \footnote[2]{llava-hf/llava-v1.6-mistral-7b-hf} \cite{liu2024llavanext}, Llama Vision \footnote[3]{meta-llama/Llama-3.2-11B-Vision-Instruct} \cite{meta2024llama3.2}, and LLaVA-COT \footnote[4]{Xkev/Llama-3.2V-11B-cot} \cite{xu2025llava}. These models are tested in a zero-shot setting on the CompliVision dataset to provide a baseline for comparison, allowing us to quantify the relative strengths and weaknesses of each VLM architecture in rule-grounded compliance prediction and interpretability.

\subsubsection{Fine-tuned Benchmarking Across Training Strategies}

To assess the upper bound of model performance, we fine-tune the LLaVA model on the CompliVision training dataset using the selected prompts: the classification-focused (C1, C4) and explanation-focused (E1, E2) prompts. This setup provides a benchmark for fully supervised training, where all training samples are annotated. We compare this baseline against our active learning strategy, which fine-tunes the model using only selectively annotated samples. This comparison enables a direct evaluation of performance gains relative to the annotation cost across different training strategies.

Our fine-tuning employs LoRA. Each selected linear layer is augmented with low-rank adapters of rank (r) = 8, scaling factor ($\alpha$) = 8, and dropout = 0.1, initialised with Gaussian-distributed weights. This configuration provides a balanced trade-off between adaptation flexibility, training stability, and computational efficiency. Fine-tuning is performed using the following hyperparameters: a learning rate of $1e^-5$, batch size of 1, and gradient accumulation over 8 steps to stabilise updates under small-batch training. The model is trained for a maximum of 10 epochs, with early stopping based on the validation edit distance metric (patience = 3) to prevent overfitting. We apply 50 warmup steps to ensure stable optimisation during early training and use gradient clipping at 1.0 to maintain numerical stability.

\subsection{Evaluation Metrics}

We evaluate the performance of zero-shot and fine-tuned tests on the CompliVision dataset using the following metrics:
\begin{itemize}
    \item \textbf{Accuracy}: The proportion of correctly predicted labels (Complied, Violated, Not Applicable) across all image-rule pairs.
    
    \item \textbf{Macro F1-Score}: The unweighted average of F1-scores for each class, providing a balanced measure of performance across all compliance labels.
    
    \item \textbf{Coverage}: The proportion of image–rule pairs for which the model produces a valid prediction, assessing model reliability and robustness.

    \item \textbf{Inference Time}: The average time required to generate predictions per image-rule pair, measuring computational efficiency and practical applicability.

\end{itemize}

To properly assess data efficiency and active learning effectiveness, we propose the Percentage of Annotation Saved $P_s$ as an extra metric for our work. During each AL and/or feedback round, the following efforts are checked: 

\begin{itemize}
    \item \textbf{Cumulative number of human labels acquired $N_h$}: reflecting true annotation effort.

    \item \textbf{Number of pseudo-labels incorporated $N_p$}: representing cost-free but model-dependent supervision.

    \item \textbf{Annotation Saved $P_s = N_p / N$}: calculated as $N_p$ divided by the total number of training samples $N = N_p + N_h$, this metric represents the fraction of training samples for which human annotation was avoided through the use of model-generated pseudo-labels.
\end{itemize}

\subsection{Implementation Details}
For model training and evaluation, image–rule pairs were sampled from each domain in the CompliVision dataset and stratified split into training, validation, and test sets as follows:

\begin{itemize}
    \item \textbf{Validation set}: 100 images × 5 rules = 500 samples (per domain)
    \item \textbf{Test set}: 100 images × 5 rules = 500 samples (per domain)
    \item \textbf{Train set}: Remaining images  × 5 rules = approx. 4,000 samples (per domain)
\end{itemize}

All experiments were conducted using Google Colab with an NVIDIA A100-SXM4-40GB GPU, leveraging pre-trained VLMs for all experiments described below, including prompt engineering, zero-shot test, fully fine-tuning test, active-learning, and explanation generation. They were implemented with the same running configurations to ensure reproducibility.

\section{Results and Discussion}

\subsection{Model Performance Across Prompt Variants} 

To examine model performance through prompt design, we explore prompt templates covering task-focused, classification-focused, and explanation-focused variants as defined in Section IV-A, all of which achieve full coverage. The task-focused variants (T1–T4) showed progressive gains in both macro-F1 and accuracy as task constraints and clarity increased from a macro-F1 of 0.2058 with the classification-task version (T1) to 0.3794 with the analysis-task version (T4). Introducing structured response formats in the classification variants further improved performance, particularly when using explicit conditional phrasing (C2), which achieved up to macro-F1 of 0.5443. Incorporating ambiguity-handling mechanisms in the C3 prompt yielded consistent gains, indicating that prompting the model to reason about uncertainty enhanced reliability. The chain-of-thought prompt (C4) provided additional performance improvements, with the "think before answering" configuration  reaching the highest macro-F1 of 0.5661 among classification-focused variants. Finally, integrating explicit explanation slightly reduced the performance to 0.5655. Although explanation prompts increased inference time to about 2.5 seconds per sample, they produced more interpretable outputs and reduced ambiguity in borderline cases, supporting their adoption as the final prompt template for feedback-aligned learning. 

\begin{table}[!h]
\begin{threeparttable}
\caption{Average Prompt Performance across 3 Domains.}
\label{tab3}
\setlength\tabcolsep{0pt} 

\begin{tabular*}{\columnwidth}{@{\extracolsep{\fill}} ll ccc}
\toprule
     Prompt & ID & 
     \multicolumn{3}{c}{Avg Model Performance} \\ 
\cmidrule{3-5}
     & & Macro F1 & Accuracy & Time (s) \\
\midrule
Task-focused & T1 & 0.2058 & 0.3067 & 0.1787 \\
 & T2 & 0.2790 & 0.3913 & 0.1853 \\
 & T3 & 0.3402 & 0.4647 & 0.1920 \\
 & T4 & \textbf{0.3794} & \textbf{0.4980} & 0.1953 \\
\midrule
Classification-focused & C1 & 0.4104 & 0.4467  & 0.5633 \\
& C2 & 0.5443 & 0.6380  & 0.5813 \\
& C3 & 0.5553 & 0.6527  & 0.9420 \\
& C4 & \textbf{0.5661} & \textbf{0.6653} & 1.0080 \\
\midrule
Explanation-focused & E1 & \textbf{0.5655} & \textbf{0.6507} & 2.4987 \\
& E2 & 0.4716 & 0.5280 & 2.1107 \\

\bottomrule
\end{tabular*}

\smallskip
\scriptsize
\end{threeparttable}
\end{table}

\subsection{Model Performance Across VLM Architectures} \label{result_baselines}

Next, we evaluated four representative VLMs: LLaVA, LLaVA-Next, LLaMA Vision, and LLaVA-COT using selected prompt variants. For LLaVA, the performance increased from a macro-F1 of 0.4266 under the basic C1 setting to 0.5738 with the explanation-enhanced E1 prompt. LLaVA-Next achieved comparable peak macro-F1 at 0.5551 but exhibited higher inference time under E2 prompt. LLaMA Vision underperformed relative to the multimodal alignment of the other models, with limited coverage and weaker text-vision fusion, though E1 still yielded a moderate improvement over its baseline. In contrast, LLaVA-COT delivered the strongest results overall, reaching a macro-F1 of 0.6358 with the E1 prompt. However, it incurred the highest inference time, with about 12 seconds per sample.

\begin{table}[!h]
\begin{threeparttable}
\caption{Average VLM Performance across 3 Domains.}
\label{tab4}
\setlength\tabcolsep{0pt} 
\begin{tabular*}{\columnwidth}{@{\extracolsep{\fill}} ll cccc}
\toprule
     VLMs & ID & 
     \multicolumn{4}{c}{Avg Model Performance} \\ 
\cmidrule{3-6}
     & & Macro F1 & Accuracy & Coverage & Time (s) \\
\midrule
LLaVA & C1 & 0.4266 & 0.4640 & 1.0000 & 0.5780 \\
 & C4 & 0.5479 & \textbf{0.6593}	& 1.0000 & 0.9647 \\
& E2 & 0.4967 & 0.5620	& 1.0000 & 2.0713 \\
 & E1 & \textbf{0.5738} & 0.6567	& 1.0000 & 2.3567 \\
\midrule
LLaVA-Next & C1 & 0.4947 & 0.6253 & 0.9633 & 0.9473 \\
 & C4 & 0.4211 & 0.6507	& 1.0000 & 1.3473 \\
& E2 & \textbf{0.5551} & \textbf{0.6567}	& 1.0000 & 3.5900 \\
 & E1 & 0.4211 & 0.6513	& 1.0000 & 3.8513 \\
\midrule
LLaMA Vision & C1 & 0.3161 & 0.3267 & 0.9880 & 0.7073 \\
 & C4 & 0.4332 & 0.3427	& 0.7947 & 3.2493 \\
& E2 & 0.3445 & 0.3300	& 0.9313 & 3.6293 \\
 & E1 & \textbf{0.4341} & \textbf{0.3627}	& 0.8233 & 5.8927 \\
\midrule
LLaVA-COT & C1 & 0.5366 & 0.5520 & 1.0000 & 7.7487 \\
 & C4 & 0.6332 & \textbf{0.6827}	& 1.0000 & 10.2467 \\
 & E2 & 0.5846 & 0.6073	& 1.0000 & 11.1460 \\
 & E1 & \textbf{0.6358} & 0.6793	& 1.0000 & 12.1053 \\
\bottomrule
\end{tabular*}

\smallskip
\scriptsize
\end{threeparttable}
\end{table}

\subsection{Model Performance Across Different Training Strategies}

Table \ref{tab5} compares zero-shot, fully fine-tuned, and our developed active learning strategies for compliance classification across traffic, construction, and warehouse domains. Zero-shot performance provides a baseline, achieving moderate F1 and accuracy scores without any manual labelling. Fully fine-tuned models reach high performance but require labelling of approximately 4,000 samples per domain, representing a substantial annotation cost. 

In contrast, the active learning approach achieves comparable performance while substantially reducing manual annotation effort. For example, in the traffic domain, after four rounds of AL, the model reaches a macro-F1 score of 0.8416 and accuracy of 0.8920 using only 1,054 manually labelled samples, supplemented with 2,946 pseudo-labelled data. Similar trends are observed in construction and warehouse environments, where AL progressively selects the most informative samples for human labelling, enabling high model performance with 65–73\% fewer manual annotations compared to full fine-tuning. These results demonstrate that active learning effectively balances annotation efficiency and model accuracy, supporting scalable and cost-effective training for compliance classification across diverse safety-critical domains. Figure 5 visualises the t-SNE embeddings and their corresponding rule-compliance regions, comparing the distribution of samples in Active Learning Round 0 and Round 3.

For performance explanation feedback, Table \ref{tab5} shows that fine-tuning with a modest set of 1,500 labelled samples incorporating in-context feedback significantly improves both F1 and accuracy across all three domains compared to zero-shot models and models trained on the same number of samples without explanations. Moreover, this approach achieves performance comparable to models trained on larger datasets without explanations, demonstrating that in-context feedback with explanations can compensate for a smaller labelled dataset and enhance the model's data efficiency. Figure \ref{fig5} presents examples of hazard detection outputs across three application domains and classification types, highlighting the adaptability of the proposed approach to varied contexts. Figure \ref{fig6} demonstrates the iterative feedback process used to further refine model predictions through human interaction.

\begin{table*}[!h]
\centering
\begin{threeparttable}
\caption{Zero-Shot, Fine-tuning, and Active Learning Performance across 3 Domains.}
\label{tab5}
\setlength{\tabcolsep}{0.7\tabcolsep}
\begin{tabular*}{0.9\textwidth}{@{\extracolsep{\fill}} llllllll cc}
\toprule
     Domain & Method & AL & Num. Manual & Accum. Manual & No. Pseudo  & Training & Annotation & 
     \multicolumn{2}{c}{Avg Model Performance} \\ 
\cmidrule{9-10}
     & & Rounds & Labels & Labels & Labels & Samples & Saved (\%) & Macro F1 & Accuracy \\
\midrule
Traffic & Zero-shot & - & 0	& 0	& 0	& 0	& 0 & 0.5773 & 0.6040 \\ 
    & + Explanation & - & 0	& 0	& 0	& 0	& 0 & 0.6182 & 0.6600 \\ 
    & Fine-tuned & - & 4035 & 4035 & 0	& 4035 & 0 & 0.8368 & 0.8960 \\
    & Active Learning & 0 & 863 & 863 & 637 & 1500 & 42.47 & 0.7954 & 0.8560 \\
    & + Explanation & - & 863 & 863 & 637 & 1500 & 42.47 & \textbf{0.8428} & \textbf{0.9040} \\
    & & 1 & 126 & 989 & 1011 & 2000 & 50.55 & 0.8268 & 0.8780 \\
    & & 2 & 27 & 1016 & 1984 & 3000 & 66.13 & 0.8401 & 0.8900 \\
    & & 3 & 38 & 1054 & 2946 & 4000 & \textbf{73.65} & 0.8416 & 0.8920 \\
\midrule
Construction & Zero-shot & - & 0	& 0	& 0	& 0	& 0 & 0.3605 & 0.3980 \\ 
    & + Explanation & - & 0	& 0	& 0	& 0	& 0 & 0.4307 & 0.5200 \\ 
    & Fine-tuned & - & 3995 & 3995 & 0	& 3995 & 0 & 0.8375 & \textbf{0.9060} \\
    & Active Learning & 0 & 1008 & 1008 & 492 & 1500 & 32.80 & 0.7680 & 0.8480 \\
    & + Explanation & - & 1008 & 1008 & 492 & 1500 & 32.80 & 0.8077 & 0.8860 \\
    & & 1 & 198 & 1206 & 794 & 2000 & 39.70 & 0.7977 & 0.8680 \\
    & & 2 & 88 & 1294 & 1701 & 2995 & 56.79 & 0.8160 & 0.8840 \\
    & & 3 & 68 & 1362 & 2633 & 3995 & \textbf{65.91} & \textbf{0.8392} & 0.8980 \\

\midrule
Warehouse & Zero-shot & - & 0	& 0	& 0	& 0	& 0 & 0.3421 & 0.3900 \\ 
    & + Explanation & - & 0	& 0	& 0	& 0	& 0 & 0.4411 & 0.5060 \\ 
    & Fine-tuned & - & 4000 & 4000 & 0	& 4000 & 0 & 0.7503 & 0.8320 \\
    & Active Learning & 0 & 961 & 961 & 539 & 1500 & 35.93 & 0.7734 & 0.8180 \\
    & + Explanation & - & 961 & 961 & 539 & 1500 & 35.93 & 0.8402 & 0.8700 \\
    & & 1 & 139 & 1100 & 900 & 2000 & 45.00 & 0.8291 & 0.8520 \\
    & & 2 & 55 & 1155 & 1845 & 3000 & 61.50 & 0.8667 & 0.8880 \\
    & & 3 & 91 & 1246 & 2754 & 4000 & \textbf{68.85} & \textbf{0.8748} & \textbf{0.8960} \\
\midrule
\bottomrule
\end{tabular*}

\smallskip
\scriptsize
\end{threeparttable}
\end{table*}

\begin{figure*}[!h]
\centering
\includegraphics[width=0.8\textwidth]{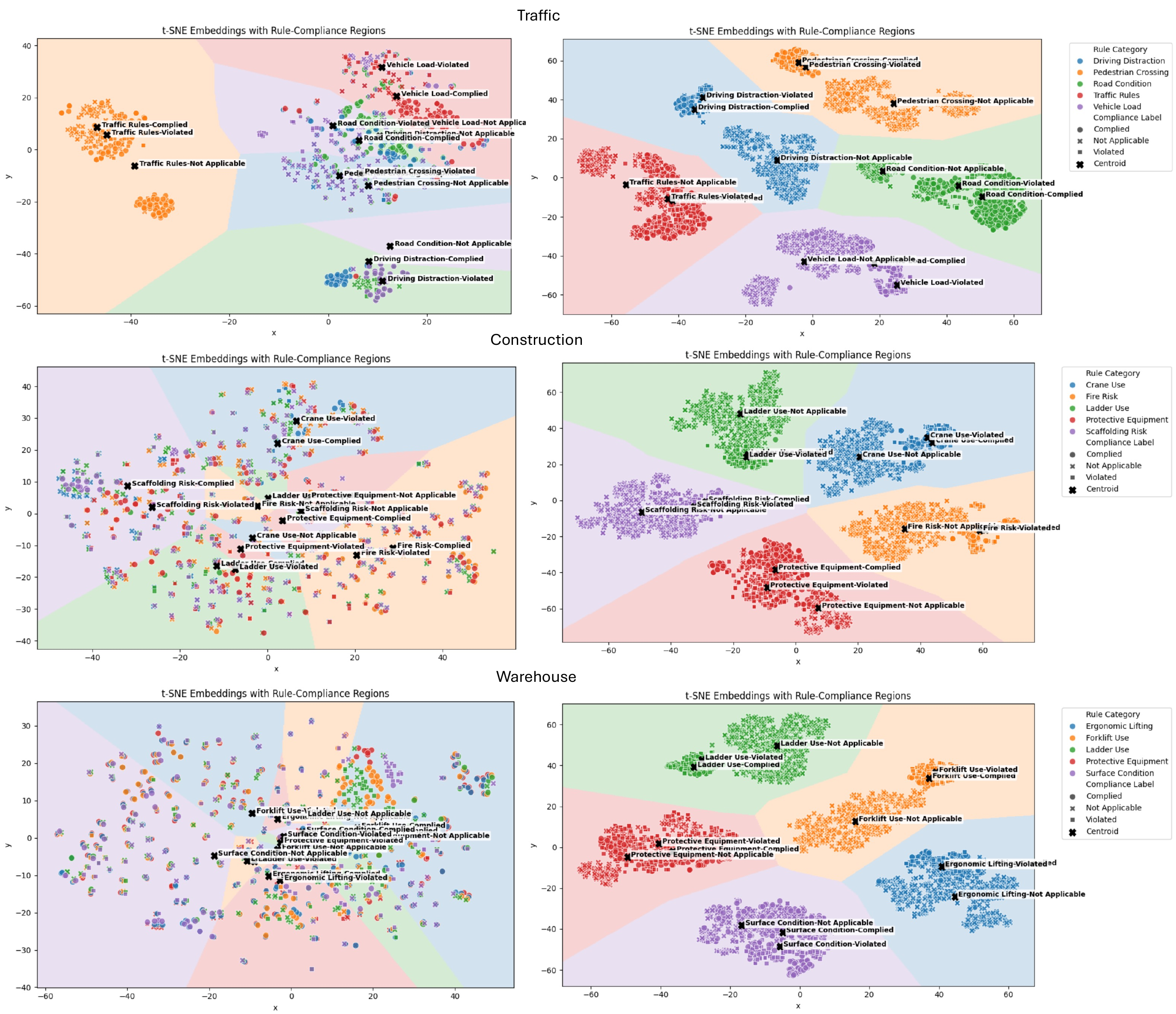}
\caption{t-SNE Embeddings with Rule-Compliance Regions in AL Round 0 vs Round 3.}
\label{fig4}
\end{figure*}

\begin{figure*}[!h]
\centering
\includegraphics[width=0.9\textwidth]{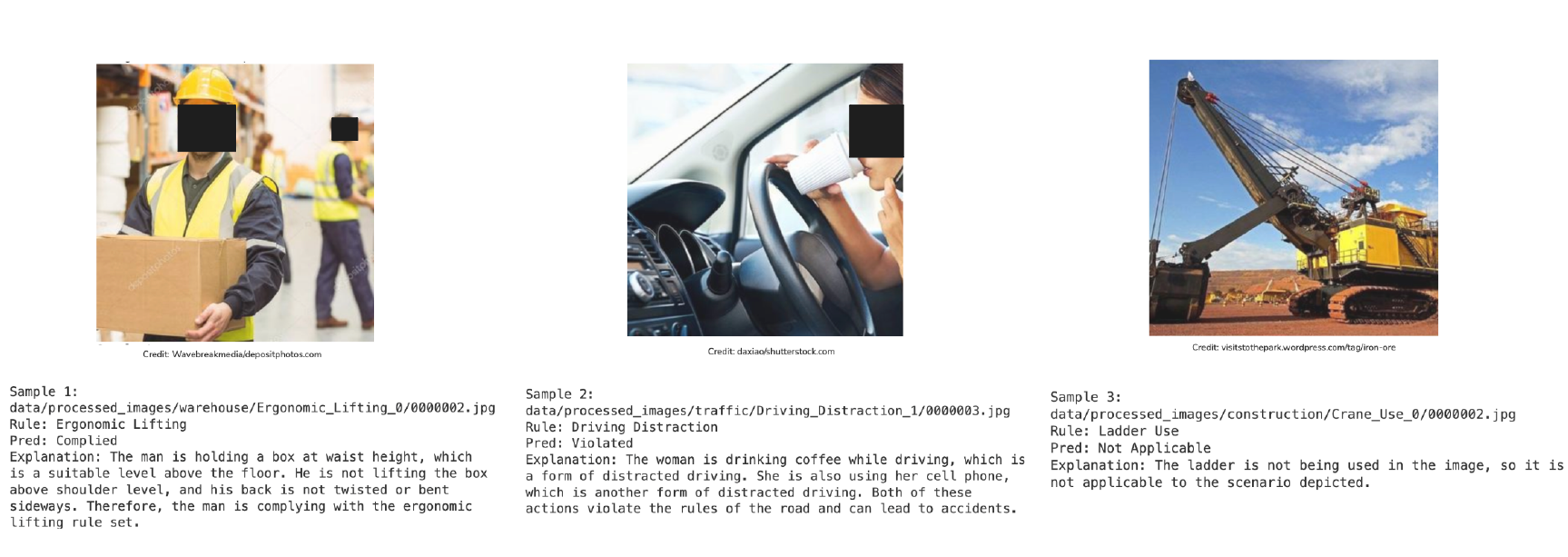}
\caption{Examples of hazard detection results across three application domains and classification types.}
\label{fig5}
\end{figure*}

\begin{figure*}[!h]
\centering
\includegraphics[width=0.9\textwidth]{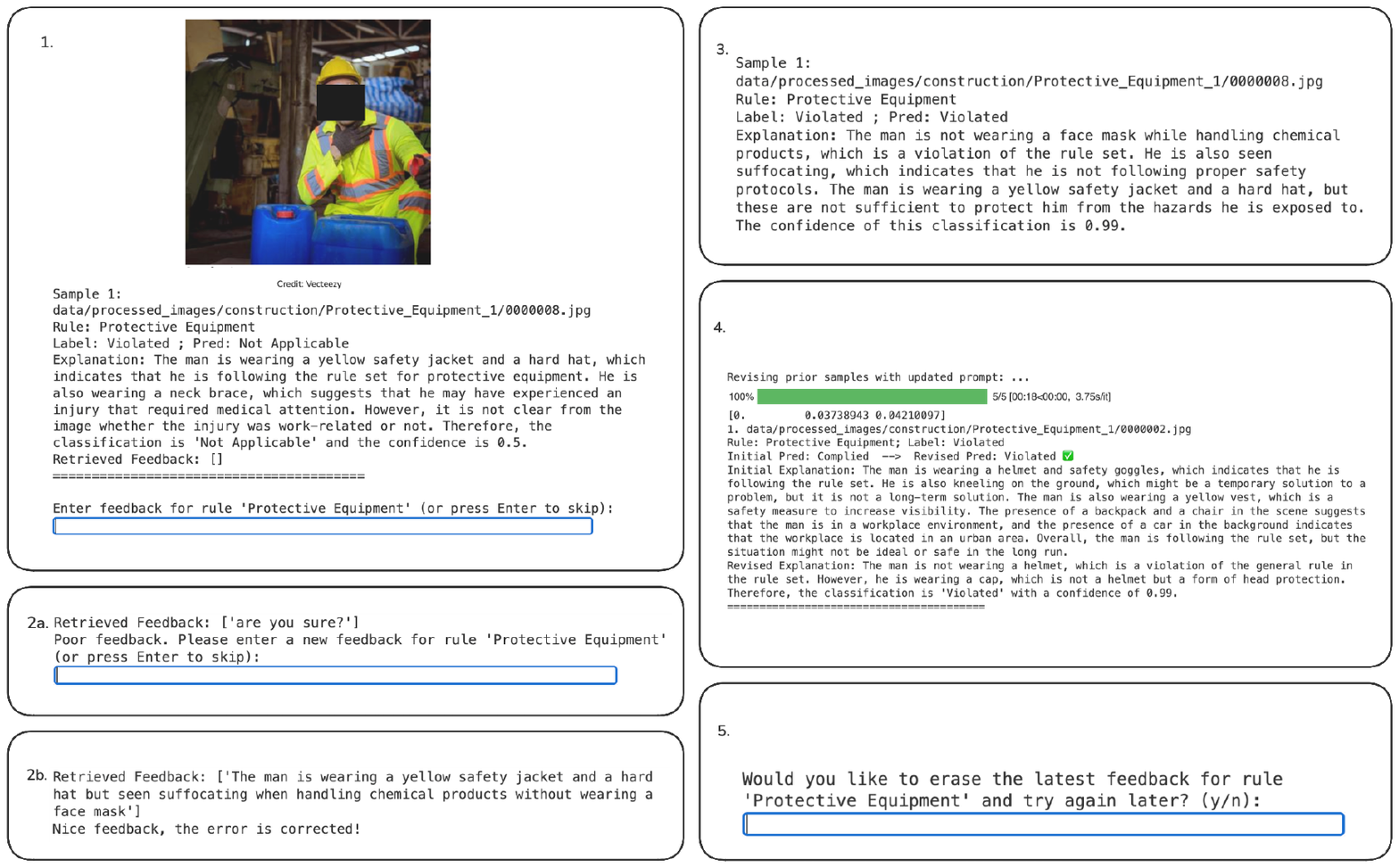}
\caption{Demonstration of the feedback process: (1) the model generates a prediction and flags weak samples for human feedback. (2a) an example of poor or non-informative feedback, prompting the user to try again. (2b) an example of effective feedback is accepted. (3) The model generates a revised output conditioned on the feedback. (4) The updated prompt is evaluated on previously seen samples to assess stability. (5) The context update is either confirmed or discarded.}
\label{fig6}
\end{figure*}

\section{Conclusion}

In conclusion, this work introduces CompliVision, a vision-language dataset and framework for rule-grounded compliance assessment across traffic, construction, and warehouse environments. By explicitly pairing images with textual safety rules and per-rule compliance labels, the dataset enables models to assess whether rules are complied with, violated, or not applicable in context, addressing a gap in existing benchmarks. Our active learning and human-in-the-loop  feedback pipeline demonstrates that high-quality compliance annotations can be obtained efficiently, reducing manual labelling by over 65\% while maintaining performance comparable to full fine-tuning. Benchmarking state-of-the-art VLMs highlights both the potential and limitations of current models for fine-grained, context-sensitive safety reasoning, particularly in complex scenes with multiple interacting rules or previously unseen scenarios. A key limitation of the current work is the challenge of fully capturing temporal dynamics in safety compliance when using static images instead of video inputs. Future work will focus on expanding CompliVision to additional safety-critical domains, incorporating temporal and multi-agent interactions, and developing more advanced multimodal reasoning strategies to further enhance interpretability, generalisation, and robustness in real-world safety monitoring.

\bibliographystyle{IEEEtran}
\bibliography{refs}

\begin{thebibliography}{10}
\providecommand{\url}[1]{#1}
\csname url@samestyle\endcsname
\providecommand{\newblock}{\relax}
\providecommand{\bibinfo}[2]{#2}
\providecommand{\BIBentrySTDinterwordspacing}{\spaceskip=0pt\relax}
\providecommand{\BIBentryALTinterwordstretchfactor}{4}
\providecommand{\BIBentryALTinterwordspacing}{\spaceskip=\fontdimen2\font plus
\BIBentryALTinterwordstretchfactor\fontdimen3\font minus \fontdimen4\font\relax}
\providecommand{\BIBforeignlanguage}[2]{{%
\expandafter\ifx\csname l@#1\endcsname\relax
\typeout{** WARNING: IEEEtran.bst: No hyphenation pattern has been}%
\typeout{** loaded for the language `#1'. Using the pattern for}%
\typeout{** the default language instead.}%
\else
\language=\csname l@#1\endcsname
\fi
#2}}
\providecommand{\BIBdecl}{\relax}
\BIBdecl

\bibitem{park2017framework}
J.~Park, K.~Kim, and Y.~K. Cho, ``Framework of automated construction-safety monitoring using cloud-enabled bim and ble mobile tracking sensors,'' \emph{Journal of Construction Engineering and Management}, vol. 143, no.~2, p. 05016019, 2017.

\bibitem{yang2019inferring}
K.~Yang and C.~R. Ahn, ``Inferring workplace safety hazards from the spatial patterns of workers’ wearable data,'' \emph{Advanced Engineering Informatics}, vol.~41, p. 100924, 2019.

\bibitem{jeelani2021real}
I.~Jeelani, K.~Asadi, H.~Ramshankar, K.~Han, and A.~Albert, ``Real-time vision-based worker localization \& hazard detection for construction,'' \emph{Automation in Construction}, vol. 121, p. 103448, 2021.

\bibitem{safeworkau}
\BIBentryALTinterwordspacing
S.~W. Australia, Oct 2025. [Online]. Available: \url{https://data.safeworkaustralia.gov.au/insights/key-whs-statistics-australia/latest-release}
\BIBentrySTDinterwordspacing

\bibitem{vukicevic2024systematic}
A.~M. Vukicevic, M.~Petrovic, P.~Milosevic, A.~Peulic, K.~Jovanovic, and A.~Novakovic, ``A systematic review of computer vision-based personal protective equipment compliance in industry practice: advancements, challenges and future directions,'' \emph{Artificial Intelligence Review}, vol.~57, no.~12, p. 319, 2024.

\bibitem{chen2025vision}
Z.~Chen, H.~Chen, M.~Imani, R.~Chen, and F.~Imani, ``Vision language model for interpretable and fine-grained detection of safety compliance in diverse workplaces,'' \emph{Expert Systems with Applications}, vol. 265, p. 125769, 2025.

\bibitem{delhi2020detection}
V.~S.~K. Delhi, R.~Sankarlal, and A.~Thomas, ``Detection of personal protective equipment (ppe) compliance on construction site using computer vision based deep learning techniques,'' \emph{Frontiers in Built Environment}, vol.~6, p. 136, 2020.

\bibitem{li2022computer}
Y.~Li, H.~Wei, Z.~Han, N.~Jiang, W.~Wang, and J.~Huang, ``Computer vision-based hazard identification of construction site using visual relationship detection and ontology,'' \emph{Buildings}, vol.~12, no.~6, p. 857, 2022.

\bibitem{chharia2025safe}
A.~Chharia, T.~Ren, T.~Furuhata, and K.~Shimada, ``Safe-construct: Redefining construction safety violation recognition as 3d multi-view engagement task,'' in \emph{Proceedings of the Computer Vision and Pattern Recognition Conference}, 2025, pp. 5811--5820.

\bibitem{our}
S.~Ng, H.~Zhou, A.~Arogbonlo, C.~P. Lim, and S.~Nahavandi, ``Vision-language hazard reasoning for driver distraction and workload estimation,'' \emph{Electronics Letters}, vol.~61, no.~1, p. e70466, 2025.

\bibitem{gil2024zero}
D.~Gil and G.~Lee, ``Zero-shot monitoring of construction workers' personal protective equipment based on image captioning,'' \emph{Automation in Construction}, vol. 164, p. 105470, 2024.

\bibitem{ding2022safety}
Y.~Ding, M.~Liu, and X.~Luo, ``Safety compliance checking of construction behaviors using visual question answering,'' \emph{Automation in Construction}, vol. 144, p. 104580, 2022.

\bibitem{yunyun2021detection}
L.~Yunyun and W.~JIANG, ``Detection of wearing safety helmet for workers based on yolov4,'' in \emph{2021 International Conference on Computer Engineering and Artificial Intelligence (ICCEAI)}.\hskip 1em plus 0.5em minus 0.4em\relax IEEE, 2021, pp. 83--87.

\bibitem{kwak2023detection}
N.~Kwak and D.~Kim, ``Detection of worker’s safety helmet and mask and identification of worker using deeplearning.'' \emph{Computers, Materials \& Continua}, vol.~75, no.~1, pp. 1671--1686, 2023.

\bibitem{pena2020real}
C.~Pena-Caballero, D.~Kim, A.~Gonzalez, O.~Castellanos, A.~Cantu, and J.~Ho, ``Real-time road hazard information system,'' \emph{Infrastructures}, vol.~5, no.~9, p.~75, 2020.

\bibitem{avazov2021fire}
\BIBentryALTinterwordspacing
K.~Avazov, M.~Mukhiddinov, F.~Makhmudov, and Y.~I. Cho, ``Fire detection method in smart city environments using a deep-learning-based approach,'' \emph{Electronics}, vol.~11, no.~1, p.~73, 2021. [Online]. Available: \url{https://www.mdpi.com/2079-9292/11/1/7}
\BIBentrySTDinterwordspacing

\bibitem{chen2021meta}
W.~Chen, Z.~Gan, L.~Li, Y.~Cheng, W.~Wang, and J.~Liu, ``Meta module network for compositional visual reasoning,'' in \emph{Proceedings of the IEEE/CVF Winter Conference on Applications of Computer Vision}, 2021, pp. 655--664.

\bibitem{yi2018neural}
K.~Yi, J.~Wu, C.~Gan, A.~Torralba, P.~Kohli, and J.~Tenenbaum, ``Neural-symbolic vqa: Disentangling reasoning from vision and language understanding,'' \emph{Advances in neural information processing systems}, vol.~31, pp. 1039--1050, 2018.

\bibitem{hudson2019gqa}
D.~A. Hudson and C.~D. Manning, ``Gqa: A new dataset for real-world visual reasoning and compositional question answering,'' in \emph{Proceedings of the IEEE/CVF conference on computer vision and pattern recognition}, 2019, pp. 6700--6709.

\bibitem{johnson2017inferring}
J.~Johnson, B.~Hariharan, L.~Van Der~Maaten, J.~Hoffman, L.~Fei-Fei, C.~Lawrence~Zitnick, and R.~Girshick, ``Inferring and executing programs for visual reasoning,'' in \emph{2017 IEEE International Conference on Computer Vision (ICCV)}, 2017, pp. 3008--3017.

\bibitem{xiao2023review}
D.~Xiao, M.~Dianati, W.~G. Geiger, and R.~Woodman, ``Review of graph-based hazardous event detection methods for autonomous driving systems,'' \emph{IEEE Transactions on Intelligent Transportation Systems}, vol.~24, no.~5, pp. 4697--4715, 2023.

\bibitem{fang2020knowledge}
W.~Fang, L.~Ma, P.~E. Love, H.~Luo, L.~Ding, and A.~Zhou, ``Knowledge graph for identifying hazards on construction sites: Integrating computer vision with ontology,'' \emph{Automation in Construction}, vol. 119, p. 103310, 2020.

\bibitem{wu2023construction}
W.~Wu, Q.~Yuan, Q.~Chen, and Y.~Cao, ``Construction safety knowledge graph integrating text and image information,'' in \emph{Proceedings of the 2023 6th International Conference on Information Management and Management Science}, 2023, pp. 26--32.

\bibitem{zhong2020hazard}
B.~Zhong, X.~Pan, P.~E. Love, J.~Sun, and C.~Tao, ``Hazard analysis: A deep learning and text mining framework for accident prevention,'' \emph{Advanced Engineering Informatics}, vol.~46, p. 101152, 2020.

\bibitem{abrecht2024deep}
S.~Abrecht, A.~Hirsch, S.~Raafatnia, and M.~Woehrle, ``Deep learning safety concerns in automated driving perception,'' \emph{IEEE Transactions on Intelligent Vehicles}, 2024.

\bibitem{dalvi2025construction}
\BIBentryALTinterwordspacing
M.~Dalvi, N.~Singh, S.~Bhingarde, and K.~Chalke, ``Construction-ppe: Personal protective equipment detection dataset,'' January 2025. [Online]. Available: \url{https://docs.ultralytics.com/datasets/detect/construction-ppe/}
\BIBentrySTDinterwordspacing

\bibitem{duan2022soda}
R.~Duan, H.~Deng, M.~Tian, Y.~Deng, and J.~Lin, ``Soda: A large-scale open site object detection dataset for deep learning in construction,'' \emph{Automation in Construction}, vol. 142, p. 104499, 2022.

\bibitem{xuehui2021dataset}
A.~Xuehui, Z.~Li, L.~Zuguang, W.~Chengzhi, L.~Pengfei, and L.~Zhiwei, ``Dataset and benchmark for detecting moving objects in construction sites,'' \emph{Automation in Construction}, vol. 122, p. 103482, 2021.

\bibitem{chen2025large}
X.~Chen and Z.~Zou, ``Are large pre-trained vision language models effective construction safety inspectors?'' \emph{arXiv preprint arXiv:2508.11011}, 2025.

\bibitem{yi2019clevrer}
K.~Yi, C.~Gan, Y.~Li, P.~Kohli, J.~Wu, A.~Torralba, and J.~B. Tenenbaum, ``Clevrer: Collision events for video representation and reasoning,'' \emph{arXiv preprint arXiv:1910.01442}, 2019.

\bibitem{wang2024deepaccident}
T.~Wang, S.~Kim, J.~Wenxuan, E.~Xie, C.~Ge, J.~Chen, Z.~Li, and P.~Luo, ``Deepaccident: A motion and accident prediction benchmark for v2x autonomous driving,'' in \emph{Proceedings of the AAAI Conference on Artificial Intelligence}, vol.~38, no.~6, 2024, pp. 5599--5606.

\bibitem{kong2024wts}
Q.~Kong, Y.~Kawana, R.~Saini, A.~Kumar, J.~Pan, T.~Gu, Y.~Ozao, B.~Opra, Y.~Sato, and N.~Kobori, ``Wts: A pedestrian-centric traffic video dataset for fine-grained spatial-temporal understanding,'' in \emph{European Conference on Computer Vision}.\hskip 1em plus 0.5em minus 0.4em\relax Springer, 2024, pp. 1--18.

\bibitem{ahmad2024sh17}
H.~M. Ahmad and A.~Rahimi, ``Sh17: A dataset for human safety and personal protective equipment detection in manufacturing industry,'' \emph{arXiv preprint arXiv:2407.04590}, 2024.

\bibitem{elhanashi2025early}
A.~Elhanashi, S.~Essahraui, P.~Dini, and S.~Saponara, ``Early fire and smoke detection using deep learning: A comprehensive review of models, datasets, and challenges,'' \emph{Applied Sciences}, vol.~15, no.~18, p. 10255, 2025.

\bibitem{liu2023visual}
H.~Liu, C.~Li, Q.~Wu, and Y.~J. Lee, ``Visual instruction tuning,'' \emph{Advances in neural information processing systems}, vol.~36, pp. 34\,892--34\,916, 2023.

\bibitem{wei2022chain}
J.~Wei, X.~Wang, D.~Schuurmans, M.~Bosma, F.~Xia, E.~Chi, Q.~V. Le, D.~Zhou \emph{et~al.}, ``Chain-of-thought prompting elicits reasoning in large language models,'' \emph{Advances in neural information processing systems}, vol.~35, pp. 24\,824--24\,837, 2022.

\bibitem{xu2025llava}
G.~Xu, P.~Jin, Z.~Wu, H.~Li, Y.~Song, L.~Sun, and L.~Yuan, ``Llava-cot: Let vision language models reason step-by-step,'' in \emph{Proceedings of the IEEE/CVF International Conference on Computer Vision}, 2025, pp. 2087--2098.

\bibitem{klein2024interactive}
L.~Klein, K.~Amara, C.~T. L{\"u}th, H.~Strobelt, M.~El-Assady, and P.~F. Jaeger, ``Interactive semantic interventions for vlms: A human-in-the-loop investigation of vlm failure,'' in \emph{Neurips Safe Generative AI Workshop 2024}, 2024.

\bibitem{kandiyana2024active}
A.~Kandiyana, P.~R. Mouton, L.~O. Hall, and D.~Goldgof, ``Active prompting of vision language models for human-in-the-loop classification and explanation of microscopy images,'' in \emph{2024 IEEE 37th International Symposium on Computer-Based Medical Systems (CBMS)}.\hskip 1em plus 0.5em minus 0.4em\relax IEEE, 2024, pp. 205--212.

\bibitem{sahoo2024prompt}
P.~Sahoo, A.~K. Singh, S.~Saha, V.~Jain, S.~Mondal, and A.~Chadha, ``A systematic survey of prompt engineering in large language models,'' \emph{arXiv preprint arXiv:2402.07927}, 2024.

\bibitem{liu2024large}
H.~Liu, C.~Li, Q.~Wu, and Y.~J. Lee, ``Large vision-language models: A survey,'' \emph{arXiv preprint arXiv:2402.14082}, 2024.

\bibitem{ren2021survey}
P.~Ren, Y.~Xiao, X.~Chang, P.-Y. Huang, Z.~Li, B.~B. Gupta, and X.~Wang, ``A survey of deep active learning,'' \emph{ACM Computing Surveys}, vol.~54, no.~9, pp. 1--40, 2021.

\bibitem{hu2022lora}
E.~J. Hu, Y.~Shen, P.~Wallis, Z.~Allen-Zhu, Y.~Li, S.~Wang, L.~Wang, W.~Chen \emph{et~al.}, ``Lora: Low-rank adaptation of large language models.'' \emph{ICLR}, vol.~1, no.~2, p.~3, 2022.

\bibitem{liu2024llavanext}
H.~Liu, C.~Li, Y.~Li, B.~Li, Y.~Zhang, S.~Shen, and Y.~J. Lee, ``Llavanext: Improved reasoning, ocr, and world knowledge,'' 2024.

\bibitem{meta2024llama3.2}
\BIBentryALTinterwordspacing
{Meta AI}, ``Llama 3.2: Revolutionizing edge ai and vision with open, customizable models,'' Sep. 2024, accessed: 2025-11-29. [Online]. Available: \url{https://ai.meta.com/blog/llama-3-2-connect-2024-vision-edge-mobile-devices/}
\BIBentrySTDinterwordspacing

\end{thebibliography}

\ifCLASSOPTIONcaptionsoff
  \newpage
\fi

\begin{IEEEbiography}[{\includegraphics[width=1in,height=1.5in,clip,keepaspectratio]{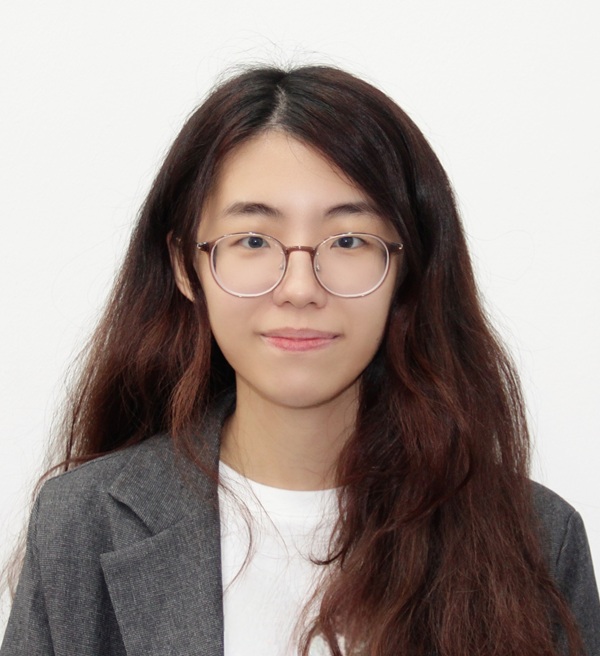}}]{Stephanie Ng} received the M.DataSc. degree from the University of Melbourne in 2021 and completed the Ph.D. degree requirements in Engineering at the Institute for Intelligent Systems Research and Innovation, Deakin University, in 2025. She is currently a Research Assistant with Deakin University and was previously a Research Associate with Swinburne University of Technology. Her research interests include generative artificial intelligence, autonomous multi-agent systems, and intelligent system design.
\end{IEEEbiography}

\begin{IEEEbiography}[{\includegraphics[width=1in,height=1.5in,clip,keepaspectratio]{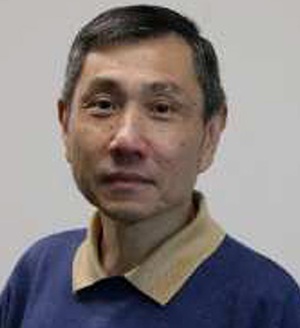}}]{CP Lim} received the Ph.D. degree from the University of Sheffield, Sheffield, U.K., in 1997. He has published over 600 scientific papers, edited eight books, and 15 invited special issues in journals, presented over a dozen of keynote talks. His research focuses on computational intelligence-based principles and their applications. Prof. Lim received ten best paper awards in international conferences.
\end{IEEEbiography}

\begin{IEEEbiography}[{\includegraphics[width=1in,height=1.5in,clip,keepaspectratio]{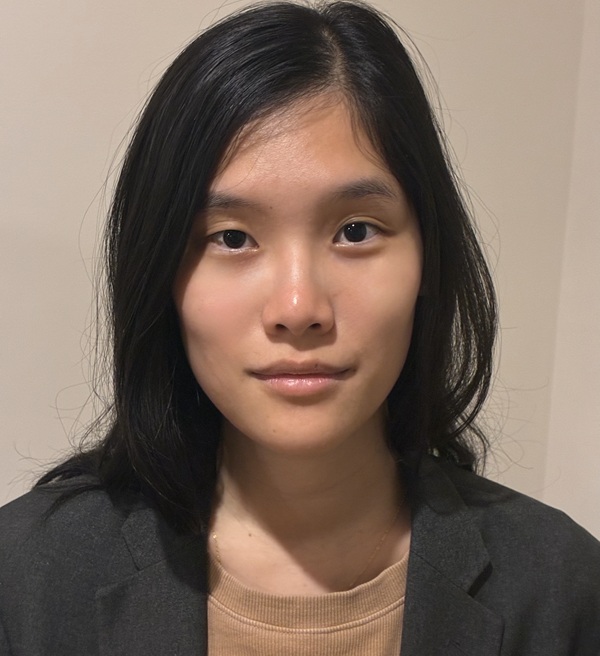}}]{SueJen Looi} received the B.Sc. degree in Applied Microbiology from Monash University, Sunway, Malaysia, in 2020, and the Master of Business Information Systems from Monash University, Clayton, Australia, in 2022. She worked as a Seasonal Research Assistant at Swinburne University. Her research interests include exploring methods and tools to enhance the performance and understanding of dynamic, data-driven systems, with applications ranging from AI models to urban infrastructure analysis.
\end{IEEEbiography}

\begin{IEEEbiography}[{\includegraphics[width=1in,height=1.5in,clip,keepaspectratio]{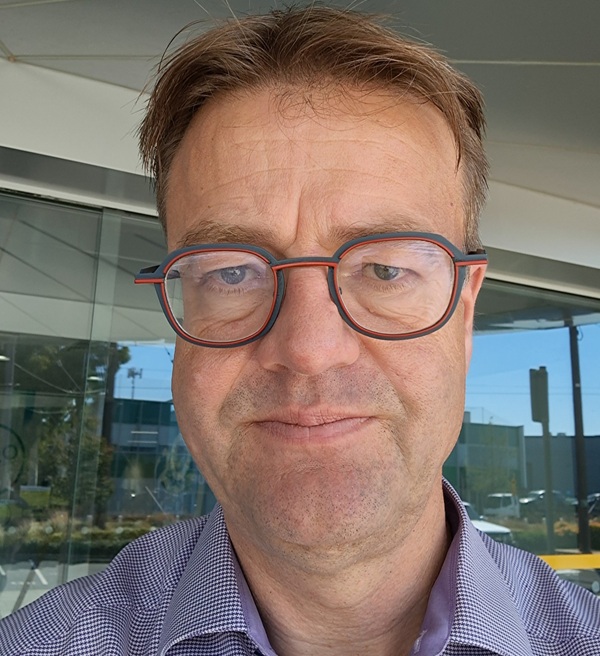}}]{Hendrik Zurlinden} received the M.Eng. degree in civil engineering from Ruhr-University, Bochum, Germany, in 1997, and the Ph.D. degree in traffic engineering from Ruhr-University, Bochum, Germany, in 2003. He is currently the Principal Engineer, Asset Performance at the Australian National Transport Research Organisation (NTRO) based in Perth, WA, Australia, where he is delivering smart motorway, road safety, and traffic management projects. Hendrik has extensive experience in advanced traffic flow theory, and he has introduced a novel concept to motorway capacity which is internationally recognised and applied today (TRB Best Paper Award 2007).
\end{IEEEbiography}

\begin{IEEEbiography}[{\includegraphics[width=1in,height=1.5in,clip,keepaspectratio]{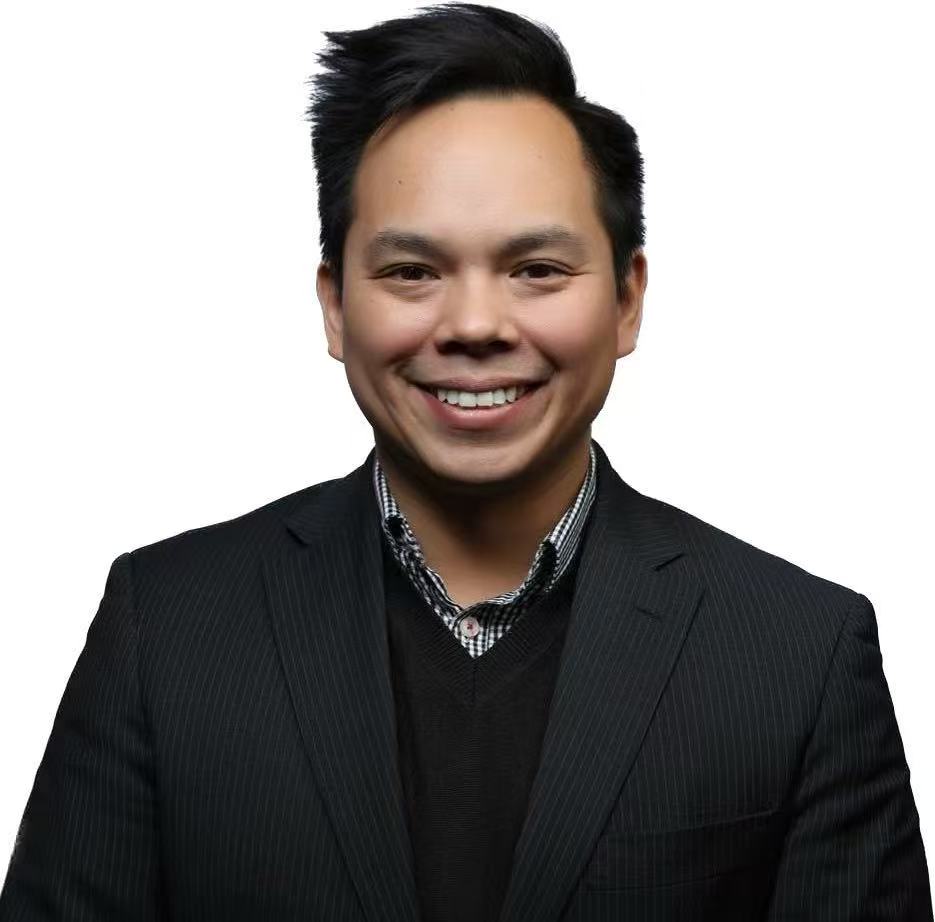}}]{David Nguyen} received a Ph.D. in Electrical Engineering and Computer Science from the University of California, Berkeley. He currently is a Technology Strategy Lead at Google. His current work at Google involves applying Generative AI approaches to solve complex challenges for global enterprises. His research focuses in areas of human-computer interaction, computer-supported collaborative work, and artificial intelligence, including topics such as computer vision and machine learning applications. 
\end{IEEEbiography}

\begin{IEEEbiography}[{\includegraphics[width=1in,height=1.5in,clip,keepaspectratio]{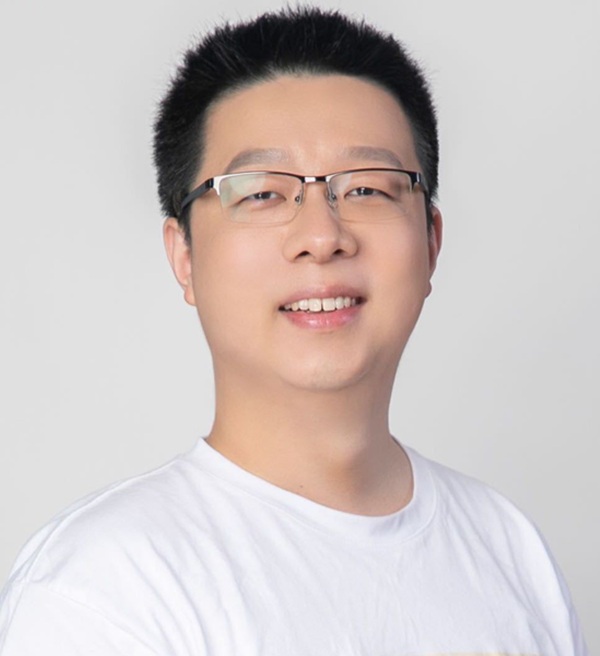}}]{Lei Wei}
received his B.Eng. degree in computer science from Tianjin University, in 2006, and the Ph.D. degree from Nanyang Technological University, Singapore, in 2011. He is currently an Associate Professor with the Intelligent Systems Research and Innovation, Deakin University, Australia. Between 2019 and 2022, Prof. Wei worked at Tencent Robotics X Lab as a Principal Research Scientist, a Project Manager, and Team Manager of the Center for Sensing, Perception and Interaction. His research interests include: haptics, HCI and VR.
\end{IEEEbiography}

\begin{IEEEbiography}[{\includegraphics[width=1in,height=1.5in,clip,keepaspectratio]{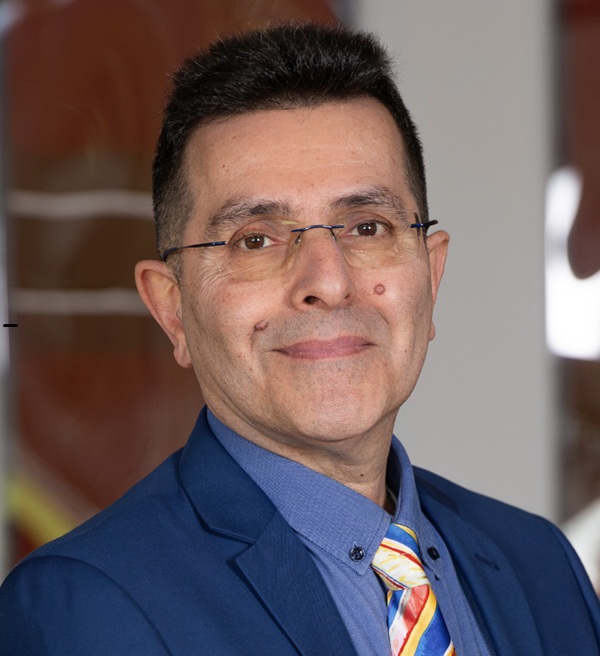}}]{Saied Nahavandi} (Fellow, IEEE) Distinguished Professor Saeid Nahavandi is Swinburne University of Technology’s inaugural Associate Deputy Vice-Chancellor Research and Chief of Defence Innovation. Saeid received a Ph.D. from Durham University, U.K. in 1991. His research interests include autonomous systems, modeling of complex systems, robotics and haptics. He has published over 1300 scientific papers in various international journals and conferences. Professor Nahavandi holds six patents, two of which have resulted in two very successful start-ups (Universal Motion Simulation Pty Ltd and FLAIM Systems Pty Ltd). Professor Nahavandi was the recipient of the Clunies Ross Entrepreneur of the Year Award 2022 from the Australian Academy of Technological Sciences \& Engineering, Researcher of the Year for Australian Space Awards 2021, Australian Defence Industry Awards - Winner of Innovator of the year, The Essington Lewis Awards, and Australian Engineering Excellence Awards - Professional Engineer of the Year.
\end{IEEEbiography}

\begin{IEEEbiography}[{\includegraphics[width=1in,height=1.5in,clip,keepaspectratio]{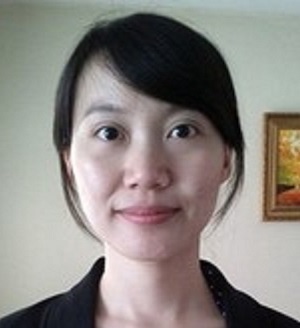}}]{Hailing Zhou} (Member, IEEE) is currently an Associate Professor in the School of Engineering, Swinburne University of Technology, Australia. She is leading research on Robotic Vision, AI, and the engineering applications in manufacturing, transportation and healthcare. She received the B.Eng. degree (Hons.) in computer science from Xidian University, China, in 2006, and the Ph.D. degree in computer vision from Nanyang Technological University (NTU), Singapore, in 2012. She subsequently joined Deakin University as a Research Fellow and later as a Senior Research Fellow. From 2020 to 2023, she worked at Accenture as an AI Principal leading innovations for industries.
\end{IEEEbiography}

\end{document}


\appendices

\section{Safety Rules} \label{appendixA}

\subsection{Traffic Domain}

\subsubsection{Driving Distraction}
\begin{itemize}
    \item Distracted driving: Distracted driving is any activity that diverts attention from driving, including talking or texting on your phone, eating and drinking, talking to people in your vehicle, fiddling with the stereo, entertainment or navigation system — anything that takes your attention away from the task of safe driving. [National Highway Traffic Safety Administration]

    \item Driver to have proper control of a vehicle etc.: A person must not drive a vehicle if a person or an animal is in the driver's lap. [ROAD SAFETY ROAD RULES 2017 - REG 297 (1A)]

    \item Touching or looking at portable devices in motor vehicles: The driver of a motor vehicle must not touch a portable device in the motor vehicle while the vehicle is moving, or is stationary but not parked. [ROAD SAFETY ROAD RULES 2017 - REG 304J (1)]

    \item Duty of driver to avoid driving while fatigued: A person must not drive a fatigue-regulated heavy vehicle on a road while the person is impaired by fatigue. [HEAVY VEHICLE NATIONAL LAW (ACT) - SECT 228 (1)]
\end{itemize}

\subsubsection{Traffic Rules}

\begin{itemize}
    \item Giving way at a pedestrian crossing: A driver must give way to any pedestrian on or entering a pedestrian crossing. [ROAD SAFETY ROAD RULES 2017 - REG 81 (2)]

    \item Overtaking or passing a vehicle at a children's crossing or pedestrian crossing: A driver approaching a children's crossing, or pedestrian crossing, must not overtake or pass a vehicle that is travelling in the same direction as the driver and is stopping, or has stopped, to give way to a pedestrian at the crossing. [ROAD SAFETY ROAD RULES 2017 - REG 82]

    \item Proceeding through a red traffic light: If traffic lights at an intersection or marked foot crossing are showing a red traffic light, a driver must not enter the intersection or marked foot crossing. [ROAD SAFETY ROAD RULES 2017 - REG 59]

    \item Driving on a one-way service road: A driver on the part of the road that is a service road must drive in the same direction as a vehicle travelling on the part of the road closest to the service road is required to travel. [ROAD SAFETY ROAD RULES 2017 - REG 136]

    \item Opening doors and getting out of a vehicle etc.: A person must not cause a hazard to any person or vehicle by opening a door of a vehicle, leaving a door of a vehicle open, or getting off, or out of, a vehicle. [ROAD SAFETY ROAD RULES 2017 - REG 269 (3)]

    \item Driving within a single marked lane or line of traffic: A driver on a multi-lane road must drive so the driver's vehicle is completely in a marked lane [ROAD SAFETY ROAD RULES 2017 - REG 146 (1)]

    \item Emergency stopping lane only signs: A driver must not drive in an emergency stopping lane. [ROAD SAFETY ROAD RULES 2017 - REG 95 (1)] 
    
    \item Stopping in an emergency stopping lane: A driver must not stop in an emergency stopping lane. [ROAD SAFETY ROAD RULES 2017 - REG 178]

    \item Parking in parking bays: A driver must position the driver's vehicle completely within a single parking bay. [ROAD SAFETY ROAD RULES 2017 - REG 211 (2)]

    \item Obstructing access to and from a footpath, driveway etc.: A driver must not stop on a road in a position that obstructs access by vehicles or pedestrians to or from a footpath ramp or a similar way of access to a footpath, or a bicycle path or passageway. [ROAD SAFETY ROAD RULES 2017 - REG 198 (1)]

\end{itemize}

\subsubsection{Pedestrian Crossing}
\begin{itemize}
    \item Crossing a road—general: A pedestrian crossing a road— (a) must cross by the shortest safe route; and (b) must not stay on the road longer than necessary to cross the road safely. [ROAD SAFETY ROAD RULES 2017 - REG 230 (1)]

    \item Crossing a road at pedestrian lights: If the pedestrian lights show a red pedestrian light and the pedestrian has not already started crossing the intersection or road, the pedestrian must not start to cross until the pedestrian lights change to green. [ROAD SAFETY ROAD RULES 2017 - REG 231 (2)]

    \item Pedestrians not to cause a traffic hazard or obstruction: A pedestrian must not cause a traffic hazard by moving into the path of a driver. [ROAD SAFETY ROAD RULES 2017 - REG 236 (1)]
\end{itemize}

\subsubsection{Road Condition}
\begin{itemize}
    \item Obligations of road users: A person who drives a motor vehicle on a highway must drive in a safe manner having regard to all the relevant factors. [ROAD SAFETY ACT 1986 - SECT 17A (1)]

    \item The relevant factors include the following— (a) the physical characteristics of the road; (b) the prevailing weather conditions; (c) the level of visibility; (d) the condition of any vehicle the person is driving or riding on the highway; (e) the prevailing traffic conditions; (f) the relevant road laws and advisory signs; (g) the physical and mental condition of the driver or road user. [ROAD SAFETY ACT 1986 - SECT 17A (2A)]

    \item Relevant vehicle not to be used in hazardous area without hazardous area authority: A person must not use a relevant vehicle in a hazardous area. [ROAD SAFETY (VEHICLES) REGULATIONS 2021 - REG 299]
\end{itemize}

\subsubsection{Vehicle Load}
\begin{itemize}
    \item Carrying goods in addition to a large indivisible item: A load-carrying vehicle must not carry more than 1 large indivisible item. [HEAVY VEHICLE (MASS, DIMENSION AND LOADING) NATIONAL REGULATION - SCHEDULE 8 Division 2 - Load-carrying vehicles 13 (1)]

    \item Load restraint requirement: The following requirements apply to a vehicle that is carrying a load— (a) the load must be secured by a means that is appropriate to the vehicle and the nature of the load; (b) the load must be placed and secured on the vehicle in a way that prevents, or would be likely to prevent, the load or any part of the load from— (i) hanging or projecting from the vehicle; or (ii) becoming dislodged or falling from the vehicle; (c) the load must not be placed or secured on the vehicle in a way that makes the vehicle unstable; (d) the load must be placed and secured on the vehicle in compliance with the performance standards recommended in the Load Restraint Guide for Light Vehicles 2018, published by the National Transport Commission. [ROAD SAFETY (VEHICLES) REGULATIONS 2021 - REG 285]
\end{itemize}

\subsection{Construction Domain}

\subsubsection{Crane Use}

\begin{itemize}
    \item The operator must not engage in any practice or activity that diverts his/her attention while actually engaged in operating the equipment, such as the use of cellular phones (other than when used for signal communications). [1926.1417(d)]

    \item Erect and maintain control lines, warning lines, railings or similar barriers to mark the boundaries of the hazard areas. [1926.1424(a)(2)(ii)]
    
    \item While the operator is not moving a suspended load, no employee must be within the fall zone [1926.1425(b)]
\end{itemize}

\subsubsection{Fire Risk}

\begin{itemize}
    \item Smoking shall be prohibited at or in the vicinity of operations which constitute a fire hazard, and shall be conspicuously posted: “No Smoking or Open Flame.” [1926.151(a)(3)]

    \item If the object to be welded, cut, or heated cannot be moved and if all the fire hazards cannot be removed, positive means shall be taken to confine the heat, sparks, and slag, and to protect the immovable fire hazards from them. [1926.352(b)]
\end{itemize}

\subsubsection{Ladder Use}

\begin{itemize}
    \item Ladders shall be used only on stable and level surfaces unless secured to prevent accidental displacement. [1926.1053(b)(6)] 
    
    \item The area around the top and bottom of ladders shall be kept clear. [1926.1053(b)(9)]
    
    \item When ascending or descending a ladder, the user shall face the ladder. [1926.1053(b)(20)]
    
    \item Each employee shall use at least one hand to grasp the ladder when progressing up and/or down the ladder. [1926.1053(b)(21)]
    
    \item An employee shall not carry any object or load that could cause the employee to lose balance and fall. [1926.1053(b)(22)]
\end{itemize}

\subsubsection{Protective Equipment}

\begin{itemize}
    \item Employees working in areas where there is a possible danger of head injury from impact, or from falling or flying objects, or from electrical shock and burns, shall be protected by protective helmets. [1926.100(a)]
    
    \item Each affected employee uses appropriate eye or face protection when exposed to eye or face hazards from flying particles, molten metal, liquid chemicals, acids or caustic liquids, chemical gases or vapors, or potentially injurious light radiation. [1926.102(a)(1)]
    
    \item Each employee on a walking/working surface (horizontal and vertical surface) with an unprotected side or edge which is 6 feet (1.8 m) or more above a lower level shall be protected from falling by the use of guardrail systems, safety net systems, or personal fall arrest systems. [1926.501(b)(1)]
\end{itemize}

\subsubsection{Scaffold Risk}

\begin{itemize}
    \item Each platform on all working levels of scaffolds shall be fully planked or decked between the front uprights and the guardrail supports [1926.451(b)(1)]
    
    \item Guardrail systems shall be installed along all open sides and ends of platforms. [1926.451(g)(4)]
    
    \item The top edge height of toprails or equivalent member on supported scaffolds shall be installed between 38 inches (0.97 m) and 45 inches (1.2 m) above the platform surface. [1926.451(g)(4)(ii)]
    
    \item In addition to wearing hardhats each employee on a scaffold shall be provided with additional protection from falling hand tools, debris, and other small objects through the installation of toeboards, screens, or guardrail systems, or through the erection of debris nets, catch platforms, or canopy structures that contain or deflect the falling objects. [1926.451(h)(1)]
\end{itemize}

\subsection{Warehouse Domain}

\subsubsection{Ergonomic Lifting}

\begin{itemize}
    \item Safe lifting involves: Holding the load close to your body at waist height. Never lift a heavy item above shoulder level. Never carry a load that obstructs your vision. [General Duty Clause, Section 5(a)(1)]
    
    \item The following points should be considered: The start and finish height of the load should be a suitable level above the floor, that is, between mid-thigh to shoulder height, preferably at about waist height. The back should not be twisted or bent sideways. Lifting with one hand should be avoided. [NOHSC:2005(1990) 5.66] 
\end{itemize}

\subsubsection{Forklift Use}

\begin{itemize}
    \item No person shall be allowed to stand or pass under the elevated portion of any truck, whether loaded or empty. [29 CFR 1910.178(m)(2)]
    
    \item All traffic regulations shall be observed, including authorized plant speed limits. A safe distance shall be maintained approximately three truck lengths from the truck ahead, and the truck shall be kept under control at all times. [1910.178(n)(1)]
    
    \item The driver shall be required to look in the direction of, and keep a clear view of the path of travel. [1910.178(n)(6)]
\end{itemize}

\subsubsection{Ladder Use}

\begin{itemize}
    \item Ladders are used only on stable and level surfaces; [29 CFR 1910.23(c)(4)]
    
    \item Each employee faces the ladder when climbing up or down it; [29 CFR 1910.23(b)(11)]
    
    \item Each employee uses at least one hand to grasp the ladder when climbing up and down it; and [29 CFR 1910.23(b)(12)]

    \item No employee carries any object or load that could cause the employee to lose balance and fall while climbing up or down the ladder. [29 CFR 1910.23(b)(13)]
\end{itemize}

\subsubsection{Protective Equipment}

\begin{itemize}
    \item Each affected employee uses appropriate eye or face protection when exposed to eye or face hazards from flying particles, molten metal, liquid chemicals, acids or caustic liquids, chemical gases or vapors, or potentially injurious light radiation [29 CFR 1910.133(a)(1)]
    
    \item Each affected employee wears a protective helmet when working in areas where there is a potential for injury to the head from falling objects. [29 CFR 1910.135(a)(1)]
    
    \item Personal fall protection systems must be worn with the attachment point of the body harness located in the center of the employee's back near shoulder level. The attachment point may be located in the pre-sternal position if the free fall distance is limited to 2 feet (0.6 m) or less. [29 CFR 1910.140(c)(22)]
\end{itemize}

\subsubsection{Surface Condition}
\begin{itemize}
    \item All places of employment, passageways, storerooms, service rooms, and walking-working surfaces are kept in a clean, orderly, and sanitary condition. [29 CFR 1910.22(a)(1)]
    
    \item The floor of each workroom is maintained in a clean and, to the extent feasible, in a dry condition. When wet processes are used, drainage must be maintained and, to the extent feasible, dry standing places, such as false floors, platforms, and mats must be provided. [29 CFR 1910.22(a)(2)]

    \item Walking-working surfaces are maintained free of hazards such as sharp or protruding objects, loose boards, corrosion, leaks, spills, snow, and ice. [29 CFR 1910.22(a)(3)]
\end{itemize}

\section{Annotation Assumptions} \label{appendixB}

\subsection{Traffic Domain}

\subsubsection{Driving Distraction}
\begin{itemize}
    \item No assumptions made.
\end{itemize}

\subsubsection{Traffic Rules}

\begin{itemize}
    \item Evaluated if the vehicle is traveling in its lane, moving in the same direction as traffic, or parked neatly in the correct orientation.
    \item Not evaluated (Not Applicable) if lane markings are not visible or the road is gravel.
    \item Vehicles traveling/parked on the emergency lane or on gravel next to the road are considered hazards.
\end{itemize}

\subsubsection{Pedestrian Crossing}
\begin{itemize}
    \item Evaluated only if both pedestrian legs and the road are visible; otherwise, Not Applicable.
\end{itemize}

\subsubsection{Road Condition}
\begin{itemize}
    \item Evaluated as long as part of the road is visible, even if blurred.
    \item Gravel roads or roads without visible lane markings are considered hazards.
    \item Vehicles not on a road (e.g., on grass) are Not Applicable.

\end{itemize}

\subsubsection{Vehicle Load}
\begin{itemize}
    \item All trucks are always evaluated.
    \item Vans and buses are evaluated only if obvious cargo is present on top or strapped to the vehicle.
    \item Vehicles with cargo are always evaluated; vehicles without cargo are Not Applicable.
\end{itemize}

\subsection{Construction Domain}

\subsubsection{Crane Use}

\begin{itemize}
    \item Assumed compliant if a crane (or part of it) is visible, unless there is a clear violation.
\end{itemize}

\subsubsection{Fire Risk}

\begin{itemize}
    \item Assumed violated if protective equipment rules are not met, even if fire is handled safely.
\end{itemize}

\subsubsection{Ladder Use}

\begin{itemize}
    \item No assumptions made.
\end{itemize}

\subsubsection{Protective Equipment}

\begin{itemize}
    \item Considered compliant if the worker/operator wears at least a helmet. Wearing only a high-visibility vest is a violation.
    \item Exceptions: 
        \begin{itemize}
            \item Firefighters, who may have different uniforms and may not require a helmet.
            \item If a smoke hazard is present (excluding cigarette smoke) and the worker lacks a breathing mask, it is considered a violation, even if a helmet is worn.
        \end{itemize}

    \item If the worker's head is not visible, label as Not Applicable.
    \item Exceptions:
        \begin{itemize}
            \item If tether, guardrail, safety net, or fall arrest systems are visible, it is considered compliant.
            \item If handling open fire (e.g., wielding flames) and the hand is ungloved, it is a violation (excludes cigarette smoke).
        \end{itemize}

\end{itemize}

\subsubsection{Scaffold Risk}

\begin{itemize}
    \item Label based on the presence of scaffolding, not necessity.
    \item Wooden frames are not considered scaffolding.
    \item If scaffolding is required but not visible, label as Not Applicable.
\end{itemize}

\subsection{Warehouse Domain}

\subsubsection{Ergonomic Lifting}

\begin{itemize}
    \item All lifted items are assumed heavy; items carried above shoulder level are violations, including when passed between two people.
    \item Picking up items do not close to waist level is a violation.
    \item Signs of back pain (holding back, grimacing) indicate a violation, even if ergonomics are correct.
\end{itemize}

\subsubsection{Forklift Use}

\begin{itemize}
    \item All accidents involving a forklift are considered hazards.
    \item Operator distraction (e.g., phone use, talking) is a violation.
    \item Evaluated if forklift and operators are present; during accidents, even vacant forklifts are considered a violation.
\end{itemize}

\subsubsection{Ladder Use}

\begin{itemize}
    \item Not always a violation if both hands are not on the ladder; assume user is stationary if carrying items.
    \item Reaching or carrying items above shoulder level on a ladder is a violation.
    \item Users must face ladder steps when climbing; facing any direction on a platform is allowed.
    \item Using non-ladders as ladders is a violation.
    \item Step ladders are considered ladders.
\end{itemize}

\subsubsection{Protective Equipment}

\begin{itemize}
    \item Workers must wear at least a safety helmet; absence is a violation even if wearing a high-visibility vest.
\end{itemize}

\subsubsection{Surface Condition}
\begin{itemize}
    \item Single boxes on the floor are violations.
    \item White backgrounds/floors are Not Applicable.
    \item Standing on improper surfaces (boxes, ladders, or other items) is a violation.
\end{itemize}

\section{Prompt Templates} \label{appendixC}

\subsection{Task-focused Variants}

\subsubsection{T1 (Inline Classification Instruction)}

\texttt{Classify the image into exactly one of "Complied", "Violated", or "Not Applicable" for compliance with the rule set.}

\subsubsection{T2 (Constrained Output Instruction)}

\texttt{Classify the image according to the rule set.}

\texttt{Respond with exactly one of: "Complied", "Violated", or "Not Applicable".}

\subsubsection{T3 (T2 – Alt Wording Instruction)}

\texttt{Classify the image against the rule set.}

\texttt{Respond with exactly one of: "Complied", "Violated", or "Not Applicable".}

\subsubsection{T4 (T3 – Analysis-focused Instruction)}

\texttt{Analyze the image against the rule set. 
Respond with exactly one of: "Complied", "Violated", or "Not Applicable".}

\subsection{Classification-focused Variants}

\subsubsection{C1 (T4 – JSON Dash List Instruction)}

\texttt{Analyze the image against the rule set.
Respond only with a JSON object containing a single key:
  - "classification": one of "Complied", "Violated", or "Not Applicable".}

\subsubsection{C2 (C1 + Explicit Conditional Instruction)}

\texttt{Analyze the image against the rule set. If the rule set clearly applies to the scenario depicted, determine whether it is "Complied" or "Violated". If the rule set does not clearly apply, return "Not Applicable".
Respond only with a JSON object containing a single key:
  - "classification": one of "Complied", "Violated", or "Not Applicable".}

\subsubsection{C3 (C2 + Ambiguity-Degree Instruction)}

\texttt{Analyze the image against the rule set. If the rule set clearly applies to the scenario depicted, determine whether it is "Complied" or "Violated". If the rule set does not clearly apply, return "Not Applicable".
Respond only with a JSON object containing a single key:
  - "classification": one of "Complied", "Violated", or "Not Applicable".
  - "confidence": a numeric value between 0 and 1 estimating both the likelihood that the classification is correct and how free it is from ambiguity.} 

\subsubsection{C4 (C3 + Implicit COT Before Answering)}

\texttt{Analyze the image against the rule set. If the rule set clearly applies to the scenario depicted, determine whether it is "Complied" or "Violated". If the rule set does not clearly apply, return "Not Applicable".
Think step-by-step before giving your final answer. Respond only with a JSON object containing a single key:
  - "classification": one of "Complied", "Violated", or "Not Applicable".
  - "confidence": a numeric value between 0 and 1 estimating both the likelihood that the classification is correct and how free it is from ambiguity.}

\subsection{Explanation-focused Variants} \label{appendixC_explanation}

\subsubsection{E1 (C4 + Explanation)}

\texttt{Analyze the image according to the rule set. If the rule set clearly applies to the scenario depicted, determine whether it is "Complied" or "Violated". If the rule set does not clearly apply, return "Not Applicable".
Think step-by-step before giving your final answer. Respond only with a JSON object containing a single key:
  - "classification": one of "Complied", "Violated", or "Not Applicable".
  - "explanation": an explanation for the classification made.
  - "confidence": a numeric value between 0 and 1 estimating both the likelihood that the classification is correct and how free it is from ambiguity.}

\subsubsection{E2 (C1 + Explanation)} 

\texttt{Analyze the image against the rule set.
Respond only with a JSON object containing a single key:
  - "classification": one of "Complied", "Violated", or "Not Applicable".
  - "explanation": an explanation for the classification made.
  - "confidence": a numeric value between 0 and 1 estimating both the likelihood that the classification is correct and how free it is from ambiguity.}
